\documentclass[times,twocolumn,final,longtitle]{elsarticle}
\usepackage[T1]{fontenc}
\usepackage[utf8]{inputenc} 

\usepackage{medima}
\usepackage{framed,multirow}
\usepackage{threeparttable}
\usepackage{booktabs}
\usepackage{amsmath}
\usepackage{captcont}

\usepackage{amssymb}
\usepackage{latexsym}

\usepackage{url}
\usepackage{xcolor}

\usepackage{hyperref}

\definecolor{newcolor}{rgb}{.8,.349,.1}

\journal{Medical Image Analysis}

\begin{document}

\verso{Bai \textit{et~al.}}

\begin{frontmatter}

\title{Beyond Benchmarks of IUGC: Rethinking Requirements of Deep Learning Method for Intrapartum Ultrasound Biometry from Fetal Ultrasound Videos}


\author[1,2]{Jieyun \snm{Bai}\fnref{fn2}\corref{cor1}}
\author[1]{Zihao \snm{Zhou}}
\author[1]{Yitong \snm{Tang}}
\author[3]{Jie \snm{Gan}}
\author[3]{Zhuonan \snm{Liang}}
\author[3]{Jianan \snm{Fan}} 
\author[4]{Lisa B.\snm{Mcguire}}
\author[5]{Jillian L. \snm{Clarke}}
\author[3]{Weidong \snm{Cai}}
\author[6]{Jacaueline \snm{Spurway}}
\author[7]{Yubo \snm{Tan}}
\author[8]{Shiye \snm{Wang}}
\author[9]{Wenda \snm{Shen}}
\author[7]{Wangwang \snm{Yu}}
\author[10]{Yihao \snm{Li}}
\author[11]{Philippe \snm{Zhang}}
\author[12]{Weili \snm{Jiang}}
\author[7]{Yongjie \snm{Li}}
\author[13]{Salem Muhsin Ali Binqahal \snm{Al Nasi}}
\author[13]{Arsen \snm{Abzhanov}}
\author[13]{Numan \snm{Saeed}}
\author[13]{Mohammad \snm{Yaqub}}
\author[14]{Zunhui \snm{Xia}}
\author[14]{Hongxing \snm{Li}}
\author[14]{Libin \snm{Lan}}
\author[15]{Jayroop \snm{Ramesh}}
\author[15]{Valentin \snm{Bacher}}
\author[16]{Mark \snm{Eid}}
\author[15]{Hoda \snm{Kalabizadeh}}
\author[16]{Christian \snm{Rupprecht}}
\author[15]{Ana I. L. \snm{Namburete}}
\author[17]{Pak-Hei \snm{Yeung}}
\author[15]{Madeleine K. \snm{Wyburd}}
\author[15]{Nicola K. \snm{Dinsdale}}
\author[18]{Assanali \snm{Serikbey}}
\author[18]{Jiankai \snm{Li}}
\author[18]{Sung-Liang \snm{Chen}}
\author[19]{Zicheng \snm{Hu}}
\author[19]{Nana \snm{Liu}}
\author[20]{Yian \snm{Deng}}
\author[19]{Wei \snm{Hu}}
\author[19]{Cong \snm{Tan}}
\author[19]{Wenfeng \snm{Zhang}}
\author[21]{Mai Tuyet \snm{Nhi}}
\author[22]{Gregor \snm{Koehler}}
\author[22]{Rapheal \snm{Stock}}
\author[22]{Klaus \snm{Maier-Hein}}
\author[23]{Marawan \snm{Elbatel}}
\author[23]{Xiaomeng \snm{Li}}
\author[24]{Saad \snm{Slimani}}
\author[25]{Victor M. \snm{Campello}\fnref{fn2}}
\author[26]{Benard \snm{Ohene-Botwe}\fnref{fn2}}
\author[35]{Isaac \snm{Khobo}\fnref{fn2}}
\author[27]{Yuxin \snm{Huang}\fnref{fn2}}
\author[28]{Zhenyan \snm{Han}\fnref{fn2}}
\author[28]{Hongying \snm{Hou}\fnref{fn2}}
\author[29]{Di \snm{Qiu}\fnref{fn2}}
\author[30]{Zheng \snm{Zheng}\fnref{fn2}\corref{cor1}}
\author[31]{Gongning \snm{Luo}\fnref{fn2}}
\author[32]{Dong \snm{Ni}\fnref{fn2}}
\author[1]{Yaosheng \snm{Lu}\fnref{fn2}\corref{cor1}}
\author[33]{Karim \snm{Lekadir}\fnref{fn2}}
\author[34]{Shuo \snm{Li}\fnref{fn2}}

\fntext[fn2]{These authors co-organized the IUGC challenge.}
\cortext[cor1]{Corresponding author: Jieyun Bai, Zheng Zheng, and Yaosheng Lu.}
\address[1]{Department of Cardiovascular Surgery, The First Affiliated Hospital of Jinan University, Jinan University, Guangzhou, China}
\address[2]{Auckland Bioengineering Institute, The University of Auckland, Auckland, New Zealand}
\address[3]{School of Computer Science, University of Sydney, Sydney, Australia}
\address[4]{Discipline of Obstetrics, Gynaecology and Neonatology, Sydney Medical School Nepean, University of Sydney Nepean Hospital, Penrith, New South Wales, Australia}
\address[5]{Discipline of Medical Imaging, Faculty of Medicine and Health, Susan Wakil Health Building, University of Sydney, Camperdown, New South Wales, Australia}
\address[6]{Medical Imaging, Orange Health Service, Orange, New South Wales, Australia}
\address[7]{University of Electronic Science and Technology of China, Chengdu, China}
\address[8]{Henan Kaifeng College of Science Technology and Communication, Kaifeng, China}
\address[9]{Changchun University of Science and Technology, Changchun, China}
\address[10]{United Imaging Healthcare, Shanghai, China}
\address[11]{University of Western Brittany, Brest, France}
\address[12]{Sichuan University, Chengdu, China}
\address[13]{Department of Machine Learning, Mohamed bin Zayed University of Artificial Intelligence, Masdar, Abu Dhabi}
\address[14]{College of Computer Science and Engineering, Chongqing University of Technology, Chongqing, China}
\address[15]{Oxford Machine Learning in NeuroImaging Lab, Department of Computer Science, University of Oxford, Oxford, United Kingdom}
\address[16]{Visual Geometry Group, University of Oxford, Oxford, United Kingdom}
\address[17]{School of Computer Science and Engineering, Nanyang Technological University, Singapore}
\address[18]{The University of Michigan-Shanghai Jiao Tong University Joint Institute, Shanghai Jiao Tong University, Shanghai, China}
\address[19]{College of Computer and Information Science, Chongqing Normal University, Chongqing, China}
\address[20]{The University of Manchester, Manchester, United Kingdom}
\address[21]{Southwest University, Chongqing, China}
\address[22]{Division of Medical Image Computing, German Cancer Research Center (DKFZ), Heidelberg, Germany}
\address[23]{Department of Electronic and Computer Engineering, The Hong Kong University of Science and Technology, Hongkong, China}
\address[24]{Chief Medical Officer Deepecho Ibn Rochd CHU, Hassan II University, Casablanca, Morocco}
\address[25]{Universitat de Barcelona, Barcelona, Spain}
\address[26]{Department of Radiography, School of Biomedical and Allied Health Sciences, College of Health Sciences, University of Ghana, Accra}
\address[35]{Department of Human Biology, Biomedical Engineering Research Center, University of Cape Town, Cape Town, South Africa}
\address[27]{Obstetrics and Gynecology Center, Zhujiang Hospital, Southern Medical University, Guangzhou, China}
\address[28]{Department of Obstetrics and Gynecology, Third Affiliated Hospital of Sun Yat-sen University, Guangzhou, China}
\address[29]{Department of Obstetrics and Gynecology, The First Affiliated Hospital of Jinan University, Guangzhou, China}
\address[30]{Guangzhou Women and Children's Medical Center, Guangdong Provincial Clinical Research Center for Child Health, Guangzhou, China}
\address[31]{Computer, Electrical and Mathematical Sciences and Engineering Division, King Abdullah University of Science and Technology, Thuwal, Saudi Arabia}
\address[32]{Shenzhen University, Shenzhen, China}
\address[33]{Artificial Intelligence in Medicine Lab (BCN-AIM), Barcelona, Spain}
\address[34]{School of Biomedical Engineering, Case Western Reserve University, Cleveland, OH, USA}

\received{30 June 2025}
\begin{abstract}
A significant proportion (45\%) of maternal deaths, neonatal deaths, and stillbirths occur during the intrapartum phase, particularly prevalent in low- and middle-income countries. Intrapartum biometry plays a crucial role in monitoring labor progress. However, the routine use of ultrasound in resource-limited settings is hindered by a shortage of trained sonographers. To tackle this issue, the Intrapartum Ultrasound Grand Challenge (IUGC), co-hosted with MICCAI 2024, was launched. The IUGC designed a multi-task automatic measurement framework oriented towards clinical applications. This framework integrates standard plane classification, fetal head-pubic symphysis segmentation, and biometry, enabling algorithms to leverage complementary information for more accurate estimations. Moreover, the challenge introduced the largest multi-center intrapartum ultrasound video dataset, consisting of 774 videos (68,106 images) collected from three hospitals. This rich dataset provides a solid foundation for algorithm training and evaluation. In this study, we elaborate on the details of the challenge, review the works submitted by eight teams, and interpret their methods from five aspects: preprocessing, data augmentation, learning strategy, model architecture, and post-processing. Additionally, we analyze the results considering various factors to identify key obstacles, explore potential solutions, and highlight ongoing challenges for future research. We conclude that although promising results have been achieved, the research remains in its early stages, and further in-depth exploration is required before clinical implementation. The solutions and the complete dataset are publicly accessible, aiming to drive continuous advancements in automatic biometry for intrapartum ultrasound imaging.
\end{abstract}


\end{frontmatter}



\section{Introduction}
\subsection{Clinical Background}
Intrapartum ultrasound has attracted substantial attention owing to its cost-effectiveness and non-invasive characteristics. It holds the potential to considerably reduce fetal mortality by as much as 20\%, presenting a remarkable prospect within the domain of maternal and fetal health \citep{grytten2018does}. In the global healthcare context, the figures related to maternal and neonatal deaths are staggering. Annually, there are 287,000 maternal deaths, 2.4 million neonatal deaths, and 1.9 million stillbirths \citep{chou2015ending}. A significant proportion, approximately 45\%, of these tragic events occur during the intrapartum phase, with a higher prevalence in low- and middle-income countries \citep{khalil2023call,ward2023simulation}. Thus, ensuring accessible and high-quality intrapartum care for all women is essential for any strategy aiming to decrease global maternal and newborn morbidity and mortality. The World Health Organization (WHO) has long underscored the importance of a woman in labor being under the vigilant monitoring of a proficient healthcare provider and the conduction of at least one ultrasound examination during each pregnancy \citep{world2018recommendations}. Additionally, organizations such as the International Society of Ultrasound in Obstetrics and Gynecology (ISUOG) \citep{ghi2018isuog}, the World Association of Perinatal Medicine (WAPM), the Perinatal Medicine Foundation (PMF) \citep{rizzo2022ultrasound}, and the National Institute for Health and Care-Excellence (NICE) \citep{blackburn2024intrapartum} have recommended intrapartum ultrasound parameters for evaluating fetal head descent prior to labor induction. These parameters are crucial prognostic tools for predicting delivery modes, as they can potentially reduce cesarean deliveries and improve the well-being of both mothers and fetuses \citep{vogel2024effects,mugyenyi2024labour}.  Therefore, the purpose of our challenge is to utilize artificial intelligence technology to achieve the automatic measurement of intrapartum ultrasound parameters. 
\subsection{Challenges}
\begin{figure*}[htbp]
	\centering
\includegraphics[width=0.7\textwidth]{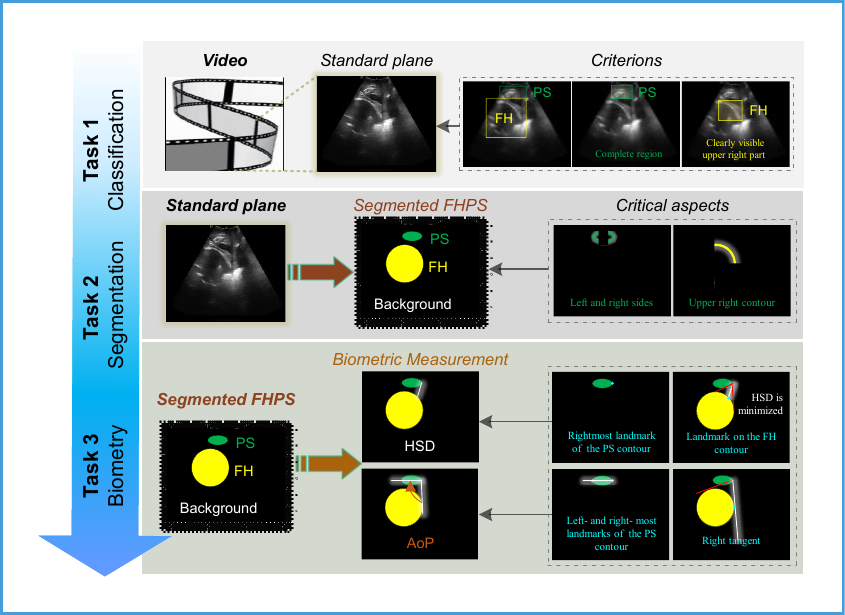}
\caption{Challenges faced by each task in the Intrapartum Ultrasound Grand Challenge (IUGC). Standard plane classification is affected by large intra-class variability caused by imaging artifacts, soft tissue deformation, fetal posture changes, and probe motion, as well as low inter-class separability due to similar echogenic patterns. Automatic segmentation of the fetal head (FH) and pubic symphysis (PS) is further complicated by labor-induced anatomical deformation, the small size of the PS, and ultrasound-specific noise, shadowing, and boundary ambiguity. Biometry estimation (AoP and HSD) requires precise geometric relationships between FH and PS, but multiple valid landmark candidates and fragmented segmentation outputs increase measurement uncertainty and algorithmic complexity.}
\label{Figure2}
\end{figure*}
Automated analysis of intrapartum ultrasound images is critically necessary \citep{fiorentino2023review,2015Standardized,rueda2013evaluation, van2019automated, slimani2023fetal,2011Standardization,zhao2023transfsm}. Manual assessment of the highly complex, rapidly changing fetal and maternal morphology in intrapartum ultrasound videos is labor-intensive, fatiguing, and prone to errors \citep{ohagwu2015intra}. The manual process involves obstetricians manipulating the transducer on the maternal abdomen to capture the fetus from various angles and depths. Imaging planes containing clinically significant fetal anatomical structures are defined as standard fetal scanning planes. Obstetricians then analyze fetal and maternal anatomy within these planes and use calipers to measure biometric parameters, such as the angle of progression (AoP) and head–symphysis distance (HSD) \citep{ghi2018isuog}. These parameters provide clinically pertinent information concerning fetal head descent prior to labor induction \citep{eggebo2024descent}. Furthermore, they can be employed for predicting the probability of a successful spontaneous vaginal delivery and for the diagnosis of fetal abnormalities \citep{eggebo2024descent}. Nevertheless, this process is highly labor-intensive and prone to human errors, exhibiting lower inter and intra-observer consistency. Obstetricians might also endure repetitive strain injuries as a result of the repetitive manipulations of the transducer and calipers \citep{janga2012work}. From a clinical perspective, analyzing fetal ultrasound images is challenging due to artifacts such as acoustic shadows, speckle noise, motion blurring, and indistinct boundaries—these arise from the interaction between ultrasound waves and the biological tissues of both the mother and fetus. Additional challenges encompass rapid fetal movements, obscured anatomical structures, and significant variability associated with different gestational weeks \citep{drukker2022clinical,ramirez2023use}.

From a technical vantage point, an automatic measurement method for intrapartum ultrasound parameters derived from videos confronts numerous challenges \citep{sappia2025acouslic}. This computer-based process comprises standard plane recognition, fetal and maternal structure segmentation, and parameter estimation \citep{2022No,0A,lee2023development,2025Diagnostic,2024Deep,gleed2023automatic,maraci2017framework,schilpzand2022automatic,0Deep,plotka2023deep}. The performance of methods in one task exerts an influence on the outcomes of subsequent tasks, and errors in each task can accumulate and ultimately have an impact on the measurement results \citep{guo2024mmsummary}. The challenges of each task are as follows (\textbf{Fig. \ref{Figure2}}).
\begin{itemize}
    \item \textbf{Classification Task:} Locating the standard planes from ultrasound videos poses several major issues. Firstly, the standard plane frequently displays high intraclass variations owing to image artifacts \citep{2021Automatic,2015Automatic,2015Standard,2016Real}. Secondly, the standard plane and non-standard plane typically possess a low interclass difference since many regions, such as the fetal head and pelvis, frequently exhibit a similar echogenicity appearance to the key structures \citep{2017SonoNet}. Finally, this low interclass difference occasionally renders it exceedingly difficult to identify the standard planes from ultrasound videos \citep{2017Ultrasound}, even for seasoned obstetric experts. In light of these challenges of high intraclass variation and low interclass difference in our task, simple low-level features might not be adequate to depict the complex appearance of the standard planes, thereby diminishing the classification performance \citep{baumgartner2017sononet,yasrab2022machine,van2018automated}.
    \item \textbf{Segmentation Task:} The automatic segmentation of the fetal head (FH) and pubic symphysis (PS) is technically demanding. Uterine contractions during labor give rise to substantial anatomical changes that impact image clarity and modify the spatial relationships between FH and PS \citep{Angeli2020New}. The relatively smaller size of the PS in comparison to the FH further complicates precise segmentation \citep{ou2024rtseg}, which is essential for accurate AoP measurements. The measurement of AoP entails three crucial landmarks - two associated with the PS and one with the FH - manifesting a significant dependency in this segmentation process. The intrinsic characteristics of ultrasound imaging pose additional challenges for automated methods \citep{bai2025psfhs}.
     \item \textbf{Biometry Task:} In accordance with the practical guidelines of ISUOG \citep{ghi2018isuog}, AoP refers to the angle between the long axis of the PS bone and a tangent line from the lowest edge of the PS to the deepest bony part of the FH, whereas HSD denotes the distance between the lowest edge of the PS and the nearest point of the FH along a line perpendicular to the long axis of the PS. A prerequisite for these measurements is the acquisition of only two parts representing the FH and PS. However, the segmented results might encompass multiple areas, rendering subsequent post-processing and measurements arduous. Furthermore, landmarks utilized for parameter measurements are typically situated on the contour of the target area, and there could be multiple landmarks that fulfill the conditions \citep{bai2022framework}. For instance, the leftmost landmark of PS or the FH tangential landmark is not singular. These intricacies thereby augment the challenges of the biometry task. 
\end{itemize}
To the best of our knowledge, few studies have reported on biometry combining intrapartum ultrasound videos. Most existing works only segment FH and PS based on single intrapartum ultrasound images \citep{bai2025psfhs}. This gap can be attributed to three fundamental challenges: (1) the absence of standardized multi-center video benchmarks, (2) the technical complexity of modeling spatio-temporal dynamics in ultrasound sequences, and (3) cumulative error propagation in cascaded classification-segmentation-measurement pipelines. Therefore, we defined the tasks and provided ultrasound videos. 
\subsection{Motivation}
To address challenges in the automatic measurement of intrapartum
ultrasound parameters based on videos, the Intrapartum Ultrasound Grand Challenge (IUGC) was organized. Specifically, the challenge provided a multi-center dataset of 774 videos from three hospitals and was aimed to encourage the development of multi-task algorithms which could perform concurrent standard plane classification, FH-PS segmentation, and parameter measurement from ultrasound videos. Sixteen submissions were assessed prior to the deadline, with eight teams presenting their work at the conference (\textbf{Fig. \ref{Figure3}}). In this paper, we present relevant information, review the methodologies, and conduct a detailed analysis of their results. Our aim is to foster interest in research exploring standardized video benchmarking, spatio-temporal feature modeling, and error-mitigation strategies for intrapartum ultrasound biometry. 
\section{Related Work}
\subsection{Related Challenges and Benchmarks}
\begin{table*}[!ht]
\centering
\caption{Summary of previous challenges related to fetal ultrasound image analysis from MICCAI/ISBI society. BMFUI: Biometric Measurements from Fetal Ultrasound Images, A-AFMA: Automatic Amniotic Fluid Measurement and Analysis from Ultrasound Video, FUGC: Fetal Ultrasound Grand Challenge: Semi-Supervised Cervical Segmentation, ACOUSLIC-AI: Abdominal Circumference Operator-agnostic Ultrasound Measurement, PSFHS: Pubic Symphysis-Fetal Head Segmentation from Transperineal Ultrasound Images, IUGC'25: Landmark Detection Challenge for Intrapartum Ultrasound Measurement Meeting the Actual Clinical Assessment
of Labor Progress; FM-UIA: Foundation Model Challenge for Ultrasound Image Analysis, IUGC: Intrapartum Ultrasound Grand Challenge 2024.}
\fontsize{7}{8} \selectfont
\setlength{\tabcolsep}{3pt}
\resizebox{\textwidth}{!}{
\begin{tabular}{@{}lcccccc@{}}
\toprule
\multirow{2}{*}{\textbf{Challenges}} & \multirow{2}{*}{\textbf{Stage}} & \multirow{2}{*}{\textbf{Targets}} & \multicolumn{3}{c}{\textbf{Tasks}} & \multirow{2}{*}{\textbf{Data size}} \\
\cmidrule(lr){4-6}
& & & \textbf{Classification} & \textbf{Segmentation} & \textbf{Biometry} & \\
\midrule
\href{https://ieeexplore.ieee.org/abstract/document/6575204}{BMFUI} Challenge (ISBI 2012) & Antepartum & Head and femur & -- & \checkmark & \checkmark & 284 images (Testing) \\
\hline
\href{https://a-afma.grand-challenge.org/}{A-AFMA} Challenge (ISBI 2021) & Antepartum & Amniotic fluid and maternal bladder & \checkmark & -- & \checkmark & -- videos \\
\hline
\href{https://www.codabench.org/competitions/4781/}{FUGC} Challenge (ISBI 2025) & Antepartum & Cervical structure & -- & \checkmark & -- & 890 images \\
\hline
\href{https://acouslic-ai.grand-challenge.org/}{ACOUSLIC-AI} Challenge (MICCAI 2024) & Antepartum & Fetal abdominal structure & \checkmark & \checkmark & -- & 582 images \\
\hline
\href{https://ps-fh-aop-2023.grand-challenge.org/}{PSFHS} Challenge (MICCAI 2023) & Intrapartum & Fetal head and pubic symphysis & -- & \checkmark & \checkmark & 5101 images \\
\hline
\href{https://www.codabench.org/competitions/7108/}{IUGC'25} Challenge (MICCAI 2025) & Intrapartum & Fetal head and pubic symphysis & -- & -- & \checkmark & 32,022 images \\
\hline
\href{https://www.codabench.org/competitions/11539/}{FM-UIA} Challenge (ISBI 2026) & Both & Maternal and fetal structures & \checkmark & \checkmark & \checkmark & 70,000 images \\
\hline
\href{https://codalab.lisn.upsaclay.fr/competitions/18413/} {IUGC} Challenge (MICCAI 2024) & Intrapartum & Fetal head and pubic symphysis & \checkmark & \checkmark & \checkmark & 774/68,106 videos/images \\
\bottomrule
\label{tab:Table1}
\end{tabular}
}
\end{table*}
\begin{table*}[!ht]
\centering
\caption{Summary of current intrapartum biometric measurement algorithms. TMP: Threshold+morphological filter+ pattern tracking methods. MT-Unet: Multi-Task Unet, DBSN: Double Branch Segmentation Network, UVSN: Ultrasound Video Segmentation Network, DSTCT: Dual-Student and Teacher Combining CNN and Transformer, PSFHSP-Net: Fetal Head and Pubic Symphysis Standard Plane Network, DBRN, FH-PSSNet, SAM: Segment Anything Model, AoP-SAM: Segment Anything Model for AoP Measurement, AoP-Net: a network based on Mask-RCNN; RMSE-root mean square error, $\Delta$AoP': Difference between Automatic and Manual measurements of AoP, $\Delta$AoP: Absolute $\Delta$AoP', $\Delta$HSD': Difference between Automatic and Manual measurements of HSD, $\Delta$HSD: Absolute $\Delta$HSD',  JC: Jaccard Coefficient; DSC: Dice Similarity Coefficient, ACC: Accuracy, ASSD: Average Symmetric Surface Distance, HD: Hausdorff Distance, F1: F1-Score, Box\_AP50: Average Precision for Detection at IoU=0.5; Seg\_AP50: Average Precision for Segmentation at IoU=0.5; AUC: Area Under the Curve; and USFM: Ultrasound Foundation Model.}
\fontsize{7}{8} \selectfont
\begin{tabular}{@{}l l l l l@{}}
\toprule
\textbf{Reference} & \textbf{Data Size} & \textbf{Tasks} & \textbf{Methods} & \textbf{Results} \\
\midrule
\citep{F2017Automatic} & 95 videos & Segmentation/biometry & TMP+Ellipse fitting & RMSE=2$^\circ$27$'$ \\
\hline
\citep{Angeli2020New} & 27 videos & Segmentation/biometry & TMP+Ellipse fitting & $\Delta$AoP'=0.99$^\circ$, $\Delta$HSD'=0.970$^\circ$ \\
\hline
\citep{zhou2020automatic} & 313 images & Segmentation/biometry & VGG+Landmarks+Ellipse fitting & DSC=0.907, $\Delta$AoP=7.60$^\circ$ \\
\hline
\citep{lu2022multitask} & 1,964 images & Classification/segmentation/biometry & MT-Unet+Landmarks+Ellipse fitting & ACC=0.989, DSC=0.920, $\Delta$AoP=3.90$^\circ$ \\
\hline
\citep{bai2022framework} & 4,056 images & Segmentation/biometry & DBSN+Ellipse fitting & DSC=0.917, ASSD=7729, $\Delta$AoP=5.11$^\circ$ \\
\hline
\citep{2022OC05}  & 1045 images & Segmentation/biometry & Unet+Landmarks+Ellipse fitting & DSC=0.921, $\Delta$AoP=5.05$^\circ$ \\
\hline
\citep{ou2024rtseg} & 4,056 images & Segmentation & ConvUNeXt & DSC=0.904, ASSD=3.62 \\
\hline
\citep{Chen2024Ultrasound}  & 289 videos & Segmentation/biometry & UVSN+Ellipse fitting & DSC=0.898, ASSD=2.416, JC=0.821, $\Delta$AoP=2.66$^\circ$ \\
\hline
\citep{jiang2024intrapartum} & 5,101 images & Segmentation & DSTCT & DSC=0.893, ASSD=0.466, HD=2.740 \\
\hline
\citep{qiu2024psfhsp} & 6,224 images & Classification & PSFHSP-Net & ACC=0.900, F1=0.908, AUC=0.933 \\
\hline
\citep{chen2024direction}  & 4700 images & Segmentation/biometry & DBRN+Ellipse fitting & DSC=0.908, HD=3.396, $\Delta$AoP=6.157$^\circ$ \\
\hline
\citep{chen2024fetal}  & 5100 images & Segmentation/biometry & FH-PSSNet+ Ellipse fitting & DSC=0.904, HD=3.476, $\Delta$AoP=6.915$^\circ$ \\
\hline
\citep{zhou2025segment} & 4,700 images & Segmentation/biometry & AoP-SAM+Ellipse fitting & DSC=0.930, ASSD=3.20, HD=13.1, $\Delta$AoP=7.74$^\circ$ \\
\hline
\citep{bai2025psfhs} & 5,101 images & Segmentation/biometry & SAM+Ellipse fitting & DSC=0.927, ASSD=3.35, HD=13.2, $\Delta$AoP=8.26$^\circ$ \\
\hline
\citep{conversano2025automated} & 45 videos & Segmentation/biometry & TMP+Ellipse fitting & RMSE=0.32 cm \\
\hline
\citep{chen2024estimation} & 1045 image & Segmentation/biometry & Unet+Ellipse fitting & $\Delta$AoP=3.777, $\Delta$HSD=4.734 \\
\hline
\citep{zhou2024evaluation} & 217 videos & Segmentation/biometry & AoP-Net+Ellipse fitting & Box\_AP50=0.907, Seg\_AP50=0.908, $\Delta$AoP'=-0.656 \\
\hline
\citep{bai2026iugc} & 32,022 images & Detection/Biometry & USFM &  $\Delta$AoP=3.81$^\circ$ \\
\bottomrule
\end{tabular}
\label{tab:Table2}
\end{table*}
In recent years, numerous challenges in detection, segmentation, biometry, and computer-aided diagnosis for obstetrics and gynecology applications have emerged \citep{bai2026fugc,lee2023development,fiorentino2023review,bai2025psfhs,van2018automated,sappia2025acouslic,rueda2013evaluation}. These challenges have fostered a collaborative environment where researchers can develop, test, and compare algorithms using standardized datasets. \textbf{Table~\ref{tab:Table1}} systematically summarizes the latest challenges and public datasets in obstetric ultrasound image analysis. Notably, eight challenges organized by the MICCAI/ISBI Society were identified \citep{bai2026fugc, bai2026iugc, van2018automated,sappia2025acouslic,rueda2013evaluation,bai_2023_7861699,bai_2024_10979813,bai_2025_15172238,bai_2024_14328192}. While most prior challenges centered on antenatal ultrasound, 
\href{https://ps-fh-aop-2023.grand-challenge.org/}{PSFHS},  \href{https://www.codabench.org/competitions/7108/}{IUGC'25},  \href{https://www.codabench.org/competitions/11539/}{FM-UIA}\citep{deng2026baseline} and \href{https://codalab.lisn.upsaclay.fr/competitions/18413} {IUGC} mark a shift toward intrapartum ultrasound applications \citep{bai2025psfhs,bai2026iugc}. This transition reflects the pressing need to address unique clinical demands during labor, where real-time and accurate fetal assessment is critical for informed decision-making.

In terms of analysis targets, prior challenges were confined to either fetal structures (e.g., head, femur, abdominal cavity) \citep{van2018automated,sappia2025acouslic,rueda2013evaluation} or maternal anatomical features (e.g., cervix, amniotic fluid, maternal bladder) \citep{jiang2025semi,chen2025comt,pham2025fetal,yu2025enhancing,tran2025human,xiao2025hierarchical}. By contrast, \href{https://ps-fh-aop-2023.grand-challenge.org/}{PSFHS}, \href{https://www.codabench.org/competitions/11539/}{FM-UIA}\citep{deng2026baseline}, \href{https://www.codabench.org/competitions/7108/}{IUGC'25} and \href{https://codalab.lisn.upsaclay.fr/competitions/18413} {IUGC} integrate both the fetal head and maternal PS, a dual-structure analysis pivotal for comprehensive intrapartum assessment \citep{bai2025psfhs} (\textbf{Table~\ref{tab:Table1}}). The interactive dynamics and relative positioning of these structures encode critical information for evaluating labor progression and fetal well-being \citep{Angeli2020New}, underscoring the clinical relevance of this integrated approach.

Shifting focus to task design, most historical challenges centered on automated measurement of linear ultrasound parameters. Unlike \href{https://ieeexplore.ieee.org/abstract/document/6575204}{BMFUI}, \href{https://a-afma.grand-challenge.org/}{A-AFMA}, \href{https://www.codabench.org/competitions/4781/}{FUGC}, and \href{https://acouslic-ai.grand-challenge.org/}{ACOUSLIC-AI} \citep{van2018automated,sappia2025acouslic,rueda2013evaluation}, which emphasize automatic measurement of length and/or circumference, \href{https://ps-fh-aop-2023.grand-challenge.org/}{PSFHS}, \href{https://www.codabench.org/competitions/11539/}{FM-UIA}\citep{deng2026baseline}, \href{https://www.codabench.org/competitions/7108/}{IUGC'25} and \href{https://codalab.lisn.upsaclay.fr/competitions/18413} {IUGC} introduce the automated measurement of angular parameters (\textbf{Table~\ref{tab:Table1}}). This focus on angles, rooted in the positional relationship between FH and PS, adds a new dimensionality to assessing labor dynamics, enabling more nuanced characterization of fetal descent.

In the context of clinical application, while \href{https://acouslic-ai.grand-challenge.org/}{ACOUSLIC-AI} \citep{sappia2025acouslic} and \href{https://codalab.lisn.upsaclay.fr/competitions/18413/} {IUGC} \citep{bai_2024_10979813} both cover detection, segmentation, and measurement tasks for ultrasound parameter automation, \href{https://codalab.lisn.upsaclay.fr/competitions/18413/} {IUGC} deals with both FH and PS, increasing the complexity of ultrasound image analysis due to the need to accurately distinguish and relate these two distinct structures. Moreover, \href{https://codalab.lisn.upsaclay.fr/competitions/18413/} {IUGC} provides a larger video dataset. Notably, apart from labeled data, it offers a substantial amount of unlabeled data, mirroring real-world clinical settings where unlabeled ultrasound data is far more abundant (\textbf{Table~\ref{tab:Table1}}). Compared to previous challenges, the quantity of videos/images in IUGC is significantly greater, enabling more robust model training and a better reflection of real-world data variability, thus pushing the boundaries of fetal ultrasound image analysis towards more practical clinical implementation.

\subsection{State-of-the-art Intrapartum Biometric Measurements}
The overview of algorithms for the automatic measurement of intrapartum ultrasound parameters is shown in \textbf{Table~\ref{tab:Table2}}. As early as 2017, Conversano et al. reported a semi-automated, non-deep learning real-time algorithm for non-invasive AoP monitoring during the second stage of labor \citep{F2017Automatic}. The initial PS/FH-containing standard plane was manually identified via image targets and pixel gray levels. PS and FH in the initial image were auto-segmented as tracking templates for subsequent frames (using similarity/cross-correlation maximization). PS axis/distal end were segmented with displacement calculated; FH edge detection was initialized via its template, with rightmost point displacement measured. Finally, FH coordinates/displacements relative to the PS distal reference system yielded the AoP \citep{conversano2025automated}. Since ISUOG introduced the Practice Guidelines for Intrapartum Ultrasound in 2018 \citep{ghi2018isuog}, a variety of deep learning strategies have been developed for automatic measurement process. In 2019, our group proposed a multitask framework based on the UNet structure for concurrently locating the two landmarks of PS and segmenting FH and PS and appraised its performance on a single-centre dataset comprising 313 images from 84 pregnant women \citep{zhou2020automatic}. In 2022, they enhanced the previous framework by incorporating a CNN branch for identifying standard planes and evaluated its performance on a single-centre dataset consisting of 1964 images of 104 volunteers during labor \citep{lu2022multitask}. To enhance segmentation and measurement accuracy, they proposed double branch segmentation networks that take into account directional information \citep{chen2024direction}, boundary information \citep{chen2024fetal,chen2024dual,chen2024estimation}, and shape prior information \citep{bai2022framework, 11356783,chen2025uncertainty}. To establish a generalized segmentation benchmark model, they released the largest multi-centre dataset containing 5101 images from 1175 pregnant women and organized the \href{https://ps-fh-aop-2023.grand-challenge.org/}{PSFHS} Challenge in conjunction with MICCAI 2023 \citep{lu2022jnu,chen2024psfhs}, with the Segment Anything Model (SAM) serving as the benchmark model \citep{bai2025psfhs}. Additional enhancements, such as the modified SAM with a multi-scale branch \citep{zhou2025segment}, the mean-teacher semi-supervised learning architecture \citep{jiang2024intrapartum, luo2026dstcs}, and knowledge distillation \citep{ou2024rtseg, qiu2024psfhsp}, were integrated to enhance measurement accuracy, fully utilize unlabeled data, and achieve real-time processing, respectively. Nonetheless, these image-based analysis methods \citep{liu2025noisy, ma2025unlabeled, kun2025ssl, tang2025heatmap, rezaei2025grm, deng2025two, zhang2025pseudo, li2025progressive, krishna2025adversarially, yang2025dsnt} are considerably distant from clinical video-based multi-task applications. Hence, the development and dissemination of annotated ultrasound video datasets from multiple devices and centers are of paramount importance for advancing end-to-end automated tools in intrapartum ultrasound imaging and facilitating robust computer-aided diagnostic systems to support labor and delivery management \citep{zhou2023segmentation, zhou2024evaluation,bai2024multimodal}.

Therefore, we delineate the IUGC and expound upon the challenge organization, the submitted multitask frameworks, and a comprehensive evaluation of the challenge results based on the Biomedical Image Analysis ChallengeS (BIAS) method \citep{maier2020bias}. The objective of the IUGC was to formulate reliable, valid, and reproducible methods for analyzing intrapartum ultrasound videos of the highly complex and rapidly evolving fetal and maternal morphology during labor. The IUGC devised multitask methods encompassing classification, segmentation, and biometry. Our evaluation compares and dissects the algorithms on a test dataset concealed from the participants. The submitted algorithms are also examined on diverse subsets of the testing dataset to ascertain whether they perform more favorably or less so under assorted circumstances, such as different data sources \citep{bai_2023_7861699,bai_2024_10979813,bai_2025_15172238}. Through this global benchmarking endeavor, we strive to optimize the multitask framework for quantitative diagnosis, proffering crucial insights into efficacious strategies and techniques. These approaches might potentially extend to other realms of video-based multitask learning and end-to-end biometrical parameter measurement, exerting a substantial influence on the broader imaging community.
\section{Materials and Setup}
\subsection{The IUGC Challenge}
The IUGC represents a collaborative initiative that encompasses the "Deep Learning in Intrapartum Ultrasound Image Analysis" cooperative group and several renowned clinical societies, namely ISUOG, WAPM, PMF, and NICE. This IUGC was hosted in conjunction with \href{https://conferences.miccai.org/2024/en/challenges.asp}{MICCAI 2024}, with the intention of facilitating the comparison of diverse algorithms within the domain of intrapartum ultrasound video analysis. It is a recurrent event with an annual submission deadline corresponding to the MICCAI schedule. The challenge was directed towards the multitask framework for conducting ultrasound parameter measurements from ultrasound videos by employing deep learning techniques.

\subsubsection{Organization}
The organizing team comprised multidisciplinary experts, including technical engineers, early-stage researchers, individuals with a PhD, assistant professors, associate professors, and full professors, in addition to practicing obstetricians and sonographers from various medical centers across Africa, Oceania, Europe, and Asia. We submitted a proposal for the IUGC challenge to the MICCAI challenge submission system, developed a \href{https://codalab.lisn.upsaclay.fr/competitions/18413}{Codalab} platform to enable participating teams to submit their methods and results, secured an \href{https://equinocs.springernature.com/home}{EquinOCS} platform for managing paper submissions related to the challenge, and—after finalizing all datasets—scheduled a timetable while designing tasks, dataset distribution, and evaluation metrics. For a detailed overview of the challenge’s setup and objectives, refer to the final challenge proposal available on \href{https://zenodo.org/records/10979813}{Zenodo} \citep{bai_2024_10979813}. Notably, while the organizers provided a baseline method for the challenge \citep{zhou2024baseline}, they were ineligible to participate.

\subsubsection{Registration and Submission}
To obtain access to the challenge dataset (available under the CC BY-NC-ND license), researchers had to sign and return a data agreement to the organizers. To support participants in engaging with the challenge, we provided a baseline method available on \href{https://github.com/maskoffs/IUGC2024}{GitHub} \citep{zhou2024baseline}. Prior to the conference, participants were assigned to develop a fully automatic multitask algorithm for intrapartum ultrasound measurements. The training dataset was made available to participants on May 15, 2024, allowing them to train their methods; they were allowed to use other publicly accessible datasets for training, provided such usage was documented in their algorithm descriptions.

\begin{figure*}[htbp]
	\centering
\includegraphics[width=0.7\textwidth]{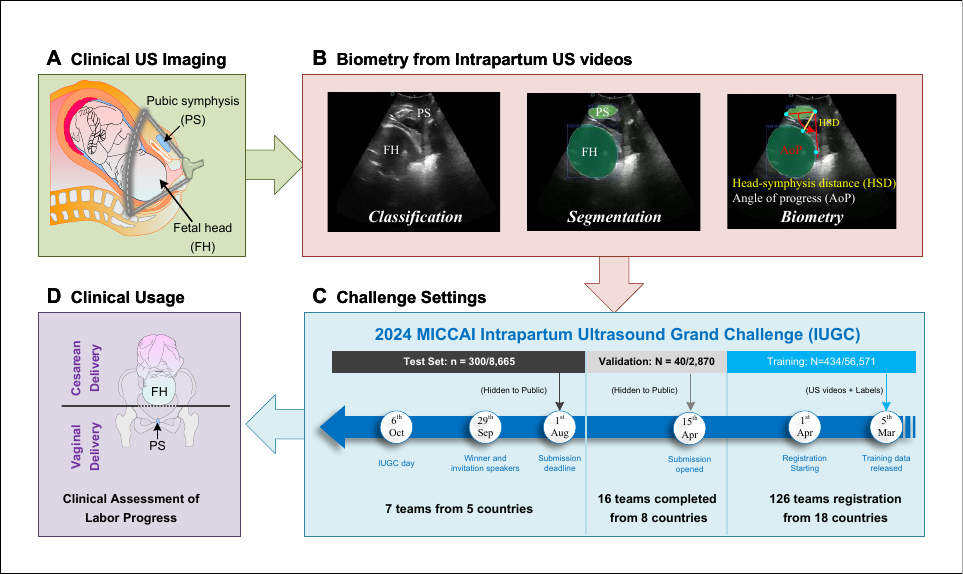}
\caption{Overall Workflow of Clinical Image Utilization and the Intrapartum Ultrasound Grand Challenge (IUGC).
A) Clinical images are acquired via a transperineal ultrasound (US) approach using mid-sagittal scans from pregnant women during labor. B) Manual operations encompass: classification of standard and non-standard planes from US videos, segmentation of the Pubic Symphysis (PS) and Fetal Head (FH) in standard plane images, and measurement of biometric parameters, namely the angle of progression (AoP) and head - symphysis distance (HSD), based on landmark annotations on the segmented results. C) In the IUGC Challenge: Dataset distribution: A total of 774 videos (68,106 images) were categorized. Specifically, 434 videos (56,571 images) were for training, 40 videos (2,870 images) for validation, and 300 videos (8,665 images) for final evaluation. The training dataset was available to all registered participants. The validation set was employed to optimize model performance during the training process, while the test set was utilized for the conclusive evaluation of these methodologies. The top eight algorithms, along with their source codes, were ranked according to classification, segmentation, and measurement metrics. D) Precise measurement of biometric parameters offers vital information for evaluating labor progression and predicting the mode of delivery.}
\label{Figure3}
\end{figure*}

Participants encapsulated their algorithms in Docker containers and submitted them to the \href{https://codalab.lisn.upsaclay.fr/competitions/18413}{Codalab} platform by August 1st, 2024. While organizers could submit containers, they were ineligible for prizes. The challenge organizers ran these containers locally on a hidden testing dataset to evaluate algorithms, allowing resubmissions only for technical issues or bugs identified during the initial assessment. The top teams received their results on September 1st, 2024, to prepare presentations, and the complete results—along with awards for the top 7 teams—were presented on October 6th, 2024, at the IUGC Challenge Session of the MICCAI (\textbf{Fig.~\ref{Figure3}}), the recording of which is available at \href{https://www.youtube.com/watch?v=xntGzr70KX8}{YouTube}. All challenge-developed algorithms were made publicly available on \href{https://github.com/maskoffs/IUGC2024}{GitHub} (\textbf{Table~\ref{tab:Table3}}).
Post-conference, the encrypted test data ground truth and evaluation tool were released, enabling subsequent participants to perform unlimited model evaluations in principle. Participants were encouraged to summarize their methods and results by submitting papers to the \href{https://equinocs.springernature.com/home}{EquinOCS} platform, adhering to LNCS style guidelines (with no page restrictions). Submissions underwent an initial review by organizers to ensure publication quality, followed by double-blind review by at least two referees, mirroring the MICCAI submission process \citep{bai2024intrapartum}.

\subsubsection{Participants}
In this challenge, a total of 126 teams from 18 countries registered, among which 16 teams from 8 countries successfully completed the competition. Finally, the top seven teams were awarded and included in the benchmark analysis, together with the official baseline model provided by the organizers. For clarity, we denote the organizer-provided baseline method as T0, while T1–T7 represent the top seven ranked teams on the final challenge leaderboard, ordered according to overall performance, with T1 corresponding to the first-place method. \textbf{Table~\ref{tab:Table3}} summarizes the participating teams, released source codes, and associated publications. In this study, we focus on a total of eight representative methods, including the organizer baseline (T0) and the top seven challenge submissions: Ganjie (T1), ViCBiC (T2), BioMedIA (T3), CQUT-Smart (T4), nkdinsdale95 (T5), HHL\_hotpot (T6), and Serikbay (T7).
\subsection{Dataset and Evaluation}
\subsubsection{Dataset}
\begin{figure*}[htbp]
	\centering
\includegraphics[width=0.7\textwidth]{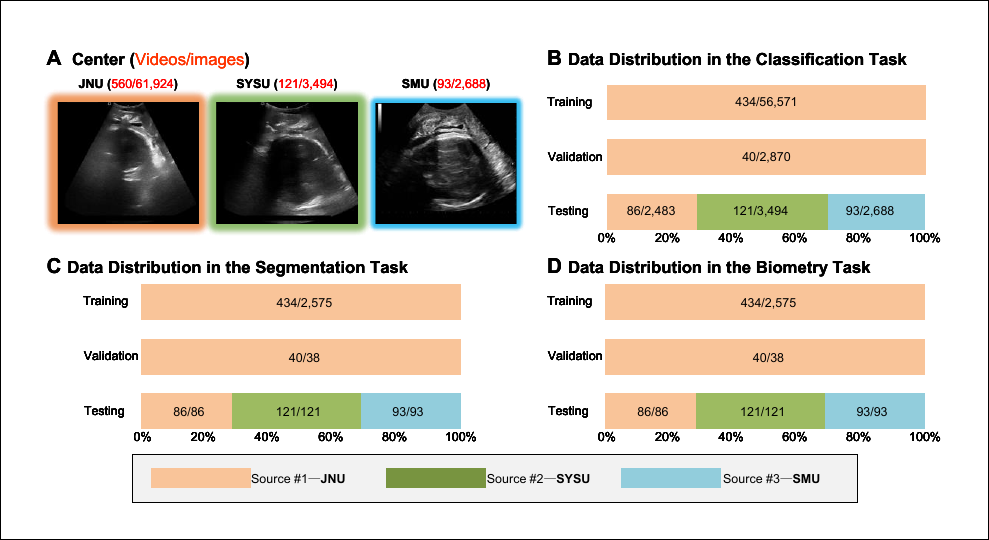}
\caption{Data sources and distribution of the ultrasound video dataset used in the IUGC challenge. A) Representative cases from each contributing hospital within the ultrasound image dataset are depicted. The First Affiliated Hospital of Jinan University (JNU) provided 560 videos with 61,924 images; the Third Affiliated Hospital of Sun Yat-sen University (SYSU) contributed 121 videos with 3,494 images; and the Zhujiang Hospital of Southern Medical University (SMU) supplied 93 videos with 2,688 images. B) Distribution of different source data across training, testing, and validation datasets (number of videos/amount of annotated data) for the classification task. C) Distribution of different source data for the segmentation task. D) Distribution of different source data for the biometric parameter measurement task.}
\label{Figure4}
\end{figure*}

The IUGC2024 dataset, used to assess fetal head station at the initiation of the second stage of labor via transperineal ultrasound, comprises 774 videos (68,106 images) retrospectively collected from 774 pregnant women (aged 18–46 years), with contributions from three medical institutions: The First Affiliated Hospital of Jinan University (JNU) provided 560 videos (61,924 images), the Third Affiliated Hospital of Sun Yat-sen University (SYSU) contributed 121 videos (3,494 images), and the Zhujiang Hospital of Southern Medical University (SMU) supplied 93 videos (2,688 images) (\textbf{Fig.~\ref{Figure4}A}). Inclusion criteria included singleton pregnancy, gestation of at least 37 + 0 weeks, and longitudinal cephalic fetal presentation, while exclusion criteria encompassed preterm delivery, non-longitudinal cephalic presentation, multiple pregnancy, uterine abnormalities, post-uterine surgery status, pathological intrapartum cardiotocography, and patient refusal. The data collected for this challenge was approved by the institutional review boards (No. JNUKY-2022-019, 2023-SYJS-023, and [2021]20-367), and informed consent was waived due to the retrospective nature of the study and the use of anonymous medical image data.

For this challenge, the dataset is partitioned into three subsets: 434 videos from JNU for training, 40 videos from JNU for validation, and 300 videos for final evaluation (comprising 83 from JNU, 121 from SYSU, and 93 from SMU). For classification tasks, the training, validation, and testing datasets consist of 434/56,571, 40/2,870, and 300/8,665 videos/annotated images, respectively (\textbf{Fig.~\ref{Figure4}B}). In segmentation and biometry tasks, the data distributions are 434/2,575/53,996, 40/38/2,832, and 300/300/8,365 (videos/annotated images/unannotated images) for training, validation, and testing, as illustrated in \textbf{Fig.~\ref{Figure4}C-D}.

All images in this study were collected by trained clinicians using three devices: Esaote MyLab (Esaote SpA, Italy), Voluson P8 (GE HealthCare, USA), and ObEye (Guangzhou Lianyin Medical Technology Co., Ltd., China), in line with protocols outlined in ISUOG guidelines \citep{ghi2018isuog}. The transducer was prepared by enclosing it in a surgical latex glove filled with coupling gel; after applying additional gel, it was positioned between the labia beneath the pubic symphysis to capture a sagittal plane. Minor lateral adjustments to the probe were made until images clearly showed maternal pelvic landmarks (pubic symphysis) and fetal landmarks (fetal skull) without shadows from the pubic rami. These videos, in avi format, were anonymized and automatically cropped (to remove headers) to 512×512 pixels before distribution. Spatial resolution (in millimeters) varied across images. By device, the distribution was as follows: the Training/Validation Set included 434/40 images from ObEye; the Testing Set contained 86 from ObEye, 121 from Voluson P8, and 93 from Esaote MyLab.

Data were acquired by specialized teams comprising sonographers, obstetricians, and technologists with over seven years of professional experience. The annotation protocol involved: 1) identifying intrapartum ultrasound standard planes by selecting high-quality images with intact PS and FH from videos; 2) free-hand manual segmentation of PS and FH; 3) measuring ultrasound parameters (i.e., AoP and HSD), where AoP denotes the angle between the long axis of the pubic bone and a tangent from the inferior margin of the pubis to the deepest bony part of the fetal skull; HSD refers to the perpendicular distance from the inferior margin of the pubic symphysis to the nearest fetal head point along the pubic axis \citep{bai_2024_10979813}. Three experts from distinct institutions annotated the data: a sonographer with ten years of ultrasound and machine learning research experience (i.e., \#1), an obstetrician with seven years of intrapartum ultrasound experience (i.e., \#2), and a sonographer with eight years of intrapartum ultrasound experience (i.e., \#3). Potential errors may arise from differences among annotators. Therefore, we randomly sampled 50 videos from three data sources, totaling 150 videos, and asked three clinicians (\#1, \#2, and \#3) to independently annotate these 150 samples. The annotations by \#1 were used as the ground truth to evaluate the accuracy and consistency of the annotations by the other two annotators. \textbf{Table~\ref{tab:Table4}} shows the results of classification, segmentation, and measurement by \#2 and \#3 relative to \#1. The results of \#2 and \#3 are highly consistent across all metrics.

\subsubsection{Evaluation Metrics}
\begin{itemize}
\item \textbf{Classification Metrics.} Four key metrics were used to evaluate model performance: Accuracy (ACC), F1-score, Matthew’s Correlation Coefficient (MCC), and the Area Under the Curve (AUC) of the receiver operating characteristic. 
\end{itemize}
\begin{itemize}
\item \textbf{Segmentation Metrics.} Three primary metrics assessed segmentation performance: Dice Similarity Coefficient (DSC), Average Symmetric Surface Distance (ASSD), and Hausdorff Distance (HD). In the study, HD corresponds to the 95th percentile Hausdorff Distance, which measures the maximum surface distance after excluding the top 5\% of outlier distances. This metric is widely adopted in medical image segmentation benchmarks to reduce sensitivity to isolated boundary noise and extreme outliers, providing a more robust estimation of contour agreement between predicted and ground-truth segmentations.
\end{itemize}
\begin{itemize}
\item \textbf{Biometry Metrics.} To gauge the accuracy of predicted ultrasound parameters (e.g., AoP, HSD), the absolute difference (denoted as $\Delta$AoP or $\Delta$HSD) between automated predictions and manual measurements was calculated for each parameter. Lower values signify higher precision in deriving clinical parameters from ultrasound images.
\end{itemize}

\subsubsection{Ranking}
We implemented a three-tiered ranking system: metric-specific ranking, task-level ranking, and challenge-wide ranking. First, models were ranked based on individual evaluation metrics. Second, task-level rankings were derived by aggregating relevant metrics for each task. Finally, challenge-wide rankings were computed by integrating all evaluation metrics across tasks. It should be noted that each metric plays an equivalent role in the ranking process.

To ensure the fairness and reproducibility of algorithm comparisons, four key analytical approaches were employed: box-and-whisker plots visualized metric distributions, highlighting minimum and maximum values, medians, 25th and 75th percentiles, and outliers; significance maps displayed results from pairwise Wilcoxon signed-rank tests, with yellow cells indicating statistically superior performance (p $<$ 0.05) of the column method over the row method, blue cells denoting comparable performance, and diagonal self-comparisons left blank; blob plots illustrated the ranking stability of each algorithm across metrics, where the median area of each blob was proportional to the relative frequency of ranks across 1,000 bootstrap samples, black crosses marked median ranks, and black lines indicated 95\% confidence intervals (spanning the 2.5th to 97.5th percentiles of the bootstrap distribution); and five methods generated rankings for each metric—MeanThenRank (calculating means before ranking), MedianThenRank (calculating medians before ranking), RankThenMean (ranking first then averaging ranks), RankThenMedian (ranking first then taking median ranks), and testBased (using Wilcoxon signed-rank tests to rank by the number of significant results). All analyses were conducted using the \href{https://github.com/wiesenfa/challengeR}{ChallengeR Toolkit}, a tool specifically designed for computing and visualizing grand challenge results \citep{wiesenfarth2021methods}.
\begin{table*}[!ht]
\centering
\caption{Summary of source code from the participants of IUGC 2024 challenge. }
\fontsize{7}{8} \selectfont
\begin{tabular}{@{}lll@{}}
\toprule
\textbf{Team} & \textbf{Code} & \textbf{Reference} \\
\midrule
Baseline (T0) & \url{https://github.com/maskoffs/IUGC-2024-Submit/tree/master/baseline} & \citep{zhou2024baseline} \\
\hline
Ganjie (T1) & \url{https://github.com/maskoffs/IUGC-2024-Submit/tree/master/ganjie} & \citep{gan2024accurate}  \\
\hline
ViCBiC (T2) & \url{https://github.com/maskoffs/IUGC-2024-Submit/tree/master/vicbic} & \citep{tan2024classification}  \\
\hline
BioMedIA (T3) & \url{https://github.com/maskoffs/IUGC-2024-Submit/tree/master/salem} & \citep{alnasi2024multi} \\
\hline
CQUT-Smart (T4) & \url{https://github.com/maskoffs/IUGC-2024-Submit/tree/master/dssaunet} & \citep{xia2024dssau}  \\
\hline
nkdinsdale95 (T5) & \url{https://github.com/maskoffs/IUGC-2024-Submit/tree/master/nkdinsdale95} & \citep{ramesh2024intrapartum}  \\
\hline
HHL\_hotpot (T6) & \url{https://github.com/maskoffs/IUGC-2024-Submit/tree/master/hhl_hotpot} & \citep{hu2024multi}  \\
\hline
Serikbay (T7) & \url{https://github.com/maskoffs/IUGC-2024-Submit/tree/master/serikbay} & \citep{serikbay2024baseline}   \\
\bottomrule
\end{tabular}
\label{tab:Table3}
\end{table*}
\begin{table*}[!ht]
\centering
\caption{The variability among the three annotations. The annotations of Annotator \#1 were used as the ground truth to evaluate the annotation results of Annotators \#2 and \#3. ACC: Accuracy; F1-score; MCC: Matthew's Correlation Coefficient; AUC: Area Under the Curve of the receiver operating characteristic; DSC\_PS/DSC\_FH: Dice Similarity Coefficient
of PS/FH; ASSD\_PS/ASSD\_FH: Average Symmetric Surface Distance of PS/FH; HD\_PS/HD\_FH: Hausdorff Distance of PS/FH; and $\Delta$AoP/$\Delta$HSD: Absolute
difference between two manual measurements of AoP/HSD. \textbf{The best results for each metric are highlighted in bold.}}
\fontsize{7}{8} \selectfont
\begin{tabular}{@{}c|cccc|cccccc|cc@{}}
\toprule
\multirow{2}{*}{\textbf{Annotator}} & \multicolumn{4}{c|}{\textbf{Classification}} & \multicolumn{6}{c|}{\textbf{Segmentation}} & \multicolumn{2}{c}{\textbf{Biometry}} \\
\cmidrule(r){2-5} \cmidrule(r){6-11} \cmidrule(l){12-13}
& {\textbf{ACC}$\uparrow$} & {\textbf{F1}$\uparrow$} & {\textbf{AUC}$\uparrow$} & {\textbf{MCC}$\uparrow$} & {\textbf{DSC\_PS}$\uparrow$} & {\textbf{DSC\_FH}$\uparrow$} & {\textbf{ASSD\_PS}$\downarrow$} & {\textbf{ASSD\_FH}$\downarrow$} & {\textbf{HD\_PS}$\downarrow$} & {\textbf{HD\_FH}$\downarrow$} & {\textbf{$\Delta$AoP}$\downarrow$} & {\textbf{$\Delta$HSD}$\downarrow$} \\
\midrule
\multirow{2}{*}{\#2} 
& \textbf{0.8707} 
& \textbf{0.7431} 
& 0.8775 
& \textbf{0.6200} 
& \textbf{0.9292} 
& \textbf{0.9614} 
& \textbf{2.7585} 
& \textbf{4.6468} 
& \textbf{9.5122} 
& \textbf{16.5011} 
& \textbf{5.0067} 
& \textbf{5.1894} \\
& (0.1546) & (0.3070) & (0.1515) & (0.3456) & (0.0560) & (0.0182) & (2.6370) & (2.3017) & (8.4864) & (7.9024) & (4.1956) & (4.5194) \\
\hline
\multirow{2}{*}{\#3} 
& 0.8649 
& 0.7332 
& \textbf{0.8788} 
& 0.5973 
& 0.9240 
& 0.9580 
& 2.9461 
& 5.0591 
& 10.2554 
& 16.8947 
& 5.2997 
& 5.6189 \\
& (0.1656) & (0.3112) & (0.1528) & (0.3710) & (0.0447) & (0.0303) & (2.0735) & (3.9696) & (8.1327) & (10.4014) & (4.2973) & (4.5845) \\
\bottomrule
\end{tabular}
\label{tab:Table4}
\end{table*}

\section{Survey of Methods}
\subsection{Description of the IUGC Methods}
The challenge included eight participating teams from diverse institutions: the Baseline (T0) team, affiliated with Jinan University (China); the Ganjie (T1) team, from the University of Sydney (Australia); the ViCBiC (T2) team, collaborating between the University of Electronic Science and Technology of China (China) and the University of Western Brittany (France); the BioMedIA (T3) team, from Mohamed bin Zayed University of Artificial Intelligence (Abu Dhabi); the CQUT-Smart (T4) team, from the College of Computer Science and Engineering, Chongqing University of Technology (China); the nkdinsdale95 (T5) team, associated with the University of Oxford (United Kingdom) and Nanyang Technological University (Singapore); the HHL\_hotpot (T6) team, from the College of Computer and Information Science, Chongqing Normal University (China); and the Serikbay (T7) team, from Shanghai Jiao Tong University (China). Each team employed distinct strategies and methodologies, which are elaborated in the following subsections.
\subsubsection{Baseline (T0)}
The baseline network is a multi-tasking model that concurrently outputs segmentation and classification results \citep{zhou2024baseline}. The segmentation branch employs a classic UNet architecture, while the classification branch leverages features extracted from the UNet's downsampling path, which are then processed by a classification head. During training, the segmentation network is optimized using a combination of Cross-Entropy (CE) Loss and Dice (DC) Loss, whereas the classification network uses only CE Loss. The training process is structured in two phases: first, the segmentation network is trained independently, followed by the classification network with the UNet parameters frozen to isolate learning. For the segmentation task, training data consists of single frames (not sequences) partitioned into a 4:1 training-to-validation split using frames with segmentation labels. Optimal segmentation weights are determined by maximizing the average DSC on the validation set for PS and FH structures. Data augmentation for segmentation includes random vertical flipping (probability 0.5), rotation ($-30^\circ$ to $+30^\circ$), and normalization of non-zero pixel regions. The AdamW optimizer is used with a learning rate of $1 \times 10^{-4}$, betas of $0.9/0.999$, weight decay of $0.1$, batch size of 4, and 80 training epochs. For the classification task, positive and negative samples are similarly split into 4:1 training/validation sets. Data augmentation here includes vertical flipping (probability 0.5) and non-zero pixel normalization. Optimal classification weights are determined by maximizing ACC on the validation set. The classification network uses identical AdamW hyperparameters but with a batch size of 8. During inference, a single frame is first classified; if deemed non-standard or unsuitable for measurement, segmentation is skipped. Otherwise, the frame proceeds through the segmentation network. 

After obtaining the segmentation results, we derive AoP and HSD through the following procedures. The measurement of HSD involves identifying two key landmarks on the segmented image: first, we locate the two farthest points on the PS contour, designating the rightmost point in the image as the first reference point for HSD. Next, using this rightmost point as a benchmark, we find the point on the PS contour that is closest to the FH contour, which serves as the second reference point. The linear distance between these two points constitutes the HSD. For AoP measurement, we construct two lines using three landmarks: starting with the two farthest points on the PS contour, we then draw a tangent line through the rightmost point of the PS to define the boundary of the FH region. The intersection of this right-side tangent line with the FH contour establishes the third landmark. The angle formed by these three points—two on the PS and one on the FH boundary—comprises the AoP. 
\subsubsection{Ganjie (T1)}
They proposed a novel approach that leverages the Video Swin Transformer to integrate local, temporal, and global features from ultrasound videos, effectively capturing comprehensive anatomical information \citep{gan2024accurate}. Additionally, they incorporated wavelet transformers within a multitask learning framework to enhance noise mitigation and improve the robustness of AoP and HSD predictions. In the classification task, given that it is a binary classification problem, they have employed the CE loss function as the primary loss function. In the segmentation task, Intersection over Union (IoU) Loss, DC Loss, and Binary CE (BCE) Loss are employed during the training process to capture the features pertinent to the segmentation task. In the biometry task, the model is required to identify critical information points for the calculation of AoP and HSD. The Mean Squared Error (MSE) loss function is deemed the biometry task. Initially, each frame was resized to a shape of (1,224,224) and the contrast of features was enhanced using histogram equalization. Before to feed the data into the Multi-tasks Spatial-Temporal Network (MTSTN), the frames were processed into video clips. They set the length of each video clip to 4 frames, balancing the trade-off between computational resources and the ability to capture dynamic features of noise in consecutive frames. They used the human position estimation to magnify the point map into the Gaussian heat-map. 
\subsubsection{ViCBiC (T2)}
To accurately classify standard planes in ultrasound videos, which include those with crucial anatomical landmarks for fetal biometry measurement, and perform precise segmentation of the FH and PS on the classified standard planes, they utilized ResNet \citep{wu2019wider} for the classification task and DeepLabV3 \citep{chen2017deeplab} for the segmentation task \citep{tan2024classification}. Since the classification task is a binary classification problem, CE loss function was employed as the primary loss function, as it effectively measures the dissimilarity between the predicted probability distribution and the true label distribution. In the segmentation task, a combination of DC Loss and BCE Loss was used during training to capture the features relevant to the task, where DC Loss emphasizes the spatial overlap between predicted and ground-truth segmentations, and BCE Loss focuses on pixel-wise classification accuracy.

The experiments were conducted on an NVIDIA TITAN RTX GPU with 24 GB of memory to provide sufficient computational resources for training these deep neural network models. The AdamW optimizer was selected with an initial learning rate of 1e-4, and a polynomial learning rate attenuation strategy was implemented to facilitate better convergence and prevent overfitting by gradually decreasing the learning rate throughout the training process. Each experiment ran for 100 epochs, and a batch size of 64 was applied consistently for both classification and segmentation tasks.

In the pre-processing stage, all images were first resized to 256×256 pixels before training. For the classification task, they were then normalized to the range of [0,1] to standardize input values and accelerate training while improving model convergence, whereas for the segmentation task, a zero-scoring normalization approach was adopted to adjust pixel values to an appropriate scale for the segmentation network. Additionally, AutoAugmentation \citep{2019Online} was employed for the classification task as a data augmentation strategy. It consists of multiple sub-strategies, with a sub-strategy randomly selected for each image in every mini-batch during training. Each sub-strategy comprises two operations, such as translation, rotation, shearing, along with specified probabilities and intensities for these operations, effectively increasing the diversity of the training data and enhancing the model's generalization ability. Finally, in the post-processing stage, test time augmentation \citep{2019Aleatoric} was implemented. The input images were horizontally and vertically flipped to generate additional test samples, increasing the quantity of test data, and the predictions from these augmented images were averaged after inference to reduce the variance of the model's predictions and improve overall prediction accuracy.
\subsubsection{BioMedIA (T3)}
Building upon the baseline method, they developed a multi-task deep learning framework with two primary objectives: 1) classifying clinically acceptable standard planes in ultrasound scans, and 2) localizing and segmenting the FH and PS to compute AoP and HSD \citep{alnasi2024multi}.

To improve training robustness, they integrated synthetic labels through pseudo-labeling and weak supervision techniques. They generated pseudo-labels using the nnUNet model, which was trained via 5-fold cross-validation on 2,575 labeled positive frames. For each fold, the training process consisted of 1,000 epochs, adopting the default 2D segmentation parameters of nnU-Net. Subsequently, they ensembled the five resulting models by averaging their softmax outputs to make inferences on unlabeled positive frames. This process yielded pseudo-labels for 22,544 images, accounting for approximately 90\% of the positive frames and 40\% of the total training set. This synthetic labeling strategy effectively expanded the training data while reducing annotation costs through weak supervision.

To address potential inaccuracies in synthetic labels, they implemented dynamic loss scaling to balance the contributions of classification and segmentation losses. The scaling factors for each loss component were calculated as the ratio of the individual loss to the total loss, resulting in a weighted sum of the classification and segmentation losses. The classification loss was computed using the standard CE loss, while the segmentation loss combined the loss from labeled data (Dice focal loss) and synthetic labeled data with an appropriate weak supervision scaling factor. Additionally, they tackled class imbalance issues by employing a weighted random sampler. This sampler undersampled negative classes and oversampled labeled and synthetic data with weights of 0.5, 0.25, and 0.25, respectively. This dual approach of loss scaling and sampling ensured balanced learning across different data types and classes.

For the optimizer, they adopted a Schedule-Free approach, which dynamically adjusted the learning rates without relying on predefined schedules. This method optimized the gradient magnitudes for each task, facilitating better convergence and balancing the performance of multiple tasks without the need for manual tuning. By adapting the learning rates according to the real-time training requirements, their framework enhanced the stability and efficiency of simultaneous classification and segmentation learning, which is crucial for avoiding suboptimal convergence in multitask scenarios.
\subsubsection{CQUT-Smart (T4)}

The experimental framework adheres to the same configuration as the baseline method, providing a consistent comparison basis for evaluating the proposed approach. In this study, they introduce the DSSAU-Net, a novel sparse self-attention network architecture specifically designed for efficient and accurate segmentation of FH and PS \citep{xia2024dssau}.

The DSSAU-Net features a symmetric U-shaped encoder-decoder structure, a well-established design in semantic segmentation tasks. At each stage of the network, different numbers of Dual Sparse Selection Attention (DSSA) blocks are stacked. The key innovation of the DSSA block resides in its capacity to execute explicit sparse token selection at both regional and pixel levels for each query. This mechanism significantly reduces computational complexity by discarding redundant computations while ensuring that the most relevant features for segmentation are retained. To tackle the problem of information loss during upsampling, convolutional skip connections are smoothly incorporated into the network architecture. These skip connections narrow the semantic gap between the encoder and decoder, facilitating the reuse of low-level features. Furthermore, multiscale feature fusion is utilized, which successfully merges global and local contextual information. For model training, a hybrid loss function that combines DC loss and CE loss is utilized. This combination ensures a balanced optimization process, promoting both precise boundary delineation and accurate class prediction.

The experiments were performed on a high-performance Geforce RTX 3090 GPU with 24 GB of memory, offering sufficient computational resources to train the deep neural network. Prior to inputting data into the model, all images are resized to a 256×256 pixel resolution. Within the DSSA blocks, each stage employs 8×8 regional partitions, which balance the capture of local context and the maintenance of computational efficiency. The encoder’s channel configuration is [96, 192, 384, 768], with feature representation capacity gradually expanding as the network deepens. The decoder starts with an initial channel dimension of 64 and gradually recovers the spatial resolution through a series of upsampling operations.

The AdamW optimizer is initialized at a learning rate of \(1 \times 10^{-4}\), capitalizing on its adaptive learning rate features and weight decay regularization to boost training stability. To enhance generalization, data augmentation methods such as rotation, flipping, and contrast adjustment are implemented, augmenting the training dataset with synthetic variations. The backbone network undergoes pre-training on the ImageNet dataset to utilize transfer learning, thereby speeding up convergence and elevating performance in medical image segmentation tasks. The sparse scaling factor \(\lambda\) is carefully set to \(1/8\), striking an optimal balance between feature density and computational efficiency to handle high-resolution medical images effectively.
\subsubsection{nkdinsdale95 (T5)}
The automated fetal biometry pipeline introduced by \cite{ramesh2024intrapartum} is a sophisticated system designed to address the challenges inherent in ultrasound image analysis. To improve the efficacy of the classification task, they utilized sparse sampling to address class imbalances and spurious correlations. For both classification and segmentation tasks, ensemble deep learning models were leveraged—an approach that ensures strong generalization across diverse ultrasound acquisition settings, which differ in imaging equipment, operator techniques, and patient characteristics. Additionally, post-processing steps are integral to the pipeline: by preserving the largest connected components and applying ellipse fitting to segmentation outputs, the structural fidelity of biometric measurements is maintained, thereby enhancing their robustness.

Data preprocessing within this pipeline is customized to the task-specific needs. For the classification task, a broad range of augmentations are applied to enhance the diversity of the training data. These augmentations include horizontal flipping, adding Gaussian noise, gamma enhancement, contrast adjustment, translation, rotation, and scaling. In contrast, for the segmentation task, affine translation and scaling are excluded to maintain spatial consistency. When flips or rotations are performed, the corresponding masks are adjusted accordingly to ensure geometric accuracy. Both the training and testing data undergo linear normalization, transforming the intensity range from \(0\)--\(255\) to \(0\)--\(1\). This normalization step is essential for compatibility with ImageNet-pre-trained models. Notably, augmentations are only applied to the training set, striking a balance between data diversity for classification and spatial integrity for segmentation.

The execution of the pipeline is divided into three distinct phases. 
\begin{enumerate}
    \item In the classification phase, models can be initialized with either random weights or weights pre-trained on ImageNet. The training process utilizes the AdamW optimizer, starting with a small initial learning rate of \(1 \times 10^{-6}\). A scheduler is implemented to halve the learning rate if the validation loss remains stagnant for 5 epochs, and early stopping is triggered if there is no improvement after 15 epochs. The loss function combines BCE with optimizer parameters \(\beta_0 = 0.9\) and \(\beta_1 = 0.999\). An ensemble of models, including EfficientNetB0, ConvNeXt, and others, achieves the optimal classification performance.
    \item In the segmentation phase, models are trained on the planes classified in the first phase. Adam optimization is used, along with a hybrid CE-Dice loss where \(\lambda_1 = \lambda_2 = 0.5\). Post-processing steps, such as retaining the major connected components, are carried out to refine the masks. The best segmentation results are obtained using an ensemble model consisting of DeepLabV3+, MA-Net, and UNet++.
    \item In the biometry phase, morphological closing with a \(10 \times 10\) pixel elliptical kernel and Canny edge-detection are first applied. Subsequently, Approximate Mean Squared ellipse fitting is used for the calculation of AoP. A decision rule is in place to prioritize either the ellipse or the hole-closed segmentation based on geometric consistency. For the measurement of HSD, ellipse fitting is bypassed, and the refined masks are used directly for landmark identification, ensuring higher accuracy.
\end{enumerate}
\subsubsection{HHL\_hotpot (T6)}
The framework inherits the baseline method’s configuration but introduces a novel automatic segmentation approach, Multi-Frequency Attention-UNeXt (MFA-UNeXt). The model employs discrete cosine transform (DCT) to decompose spatial domain information into frequency components, which are processed at both channel and spatial levels to enhance feature representation \citep{hu2024multi}. At the channel level, frequency-sensitive attention boosts sensitivity to multi-scale feature frequencies, enabling fine-grained detail extraction. At the spatial level, frequency-based operations capture both global structural context and local fine details. This dual-level attention mechanism synergistically improves the model’s capability to detect features in complex ultrasound imaging scenarios. To further enhance performance, MFA-UNeXt integrates a pyramid pooling module for multi-scale contextual aggregation and a shifted block module to expand the receptive field, thereby sharpening boundary detection and sensitivity to intricate anatomical details.

All experiments were carried out on an NVIDIA RTX 3090 GPU. The Adam optimizer was utilized with an initial learning rate of \(1 \times 10^{-4}\). Data augmentation strategies included random rotation (up to \(90^\circ\)), horizontal/vertical flipping (50\% probability each), and pixel-wise normalization, designed to enhance data diversity and model generalization. The model architecture comprises 0.26 million trainable parameters and performs 155.04 million floating-point operations (FLOPs), balancing computational efficiency with feature representation capacity. During training, the segmentation network was optimized using a hybrid loss function combining CE and DC losses, while the classification network utilized CE loss exclusively, aligning with task-specific objectives.
\subsubsection{Serikbay (T7)}
The framework retains the baseline method’s configuration while introducing tailored architectures for distinct tasks: a ResNet-50 pre-trained on ImageNet classifies standard planes in intrapartum ultrasound images, optimized for computational efficiency, and a hybrid LinkNet \citep{2017LinkNet} with a MobileNetV2 \citep{2018MobileNetV2} encoder handles segmentation tasks, balancing accuracy and resource consumption \citep{serikbay2024baseline}. The segmentation network is trained using a weighted combination of CE and DC losses to address class imbalance, while the classification network employs CE loss exclusively.

For the classification task, ResNet-50 model was initialized using ImageNet weights and trained on frames resized to 224×224, with a batch size of 64 and an initial learning rate of \(1 \times 10^{-4}\). The LinkNet segmentation model, tailored for 512×512 resolution inputs to retain anatomical details, employed a batch size of 4 and the same learning rate, adjusted to meet higher memory requirements. Both models adopted data augmentation strategies, including random horizontal/vertical flips and rotations (\(\pm 15^\circ\)), to enhance generalization.

Postprocessing refines the segmentation output through a six-step workflow: bilateral filtering smooths the segmentation while preserving edges to reduce high-frequency noise; thresholding at 0.7 converts the probability map to a binary mask, balancing sensitivity and specificity; morphological operations with a 7×7 kernel (closing and opening) remove small artifacts and fill gaps; contour extraction identifies the two largest contours from the binary mask, to which ellipses are fitted for robust anatomical structure representation; Gaussian blur softens ellipse boundaries to improve spatial consistency with original images; and mask refinement retains original pixel values within the elliptical mask while setting external regions to zero, ensuring precise anatomical localization. This structured pipeline enhances the segmentation’s alignment with anatomical shapes, minimizing false positives and boosting measurement reliability.
\subsection{Survey of the IUGC Methods}
\begin{table*}[!ht]
\centering
\caption{Summary of the benchmarked algorithms. IN: intensity normalization; SA: simple augmentation techniques, including random rotation, random flipping, random scaling, random shifting, random cropping, random warping and horizontally flipping; JNU-IFM: image dataset of the Intelligent Fetal Monitoring Lab of Jinan University \citep{lu2022jnu}; PSFHS: image dataset of pubic symphysis and fetal head segmentation challenge \citep{chen2024psfhs}; GN: Gaussian noise; GE: Gamma Enhancement; CA: Contrast Adjustment; GD: Grid distortions; CJ: Color Jitter; MC: Morphological closing; and ED: Edge-detection.}
\fontsize{7}{8} \selectfont
\begin{tabular}{@{}llllll@{}}
\toprule
\textbf{Teams} & \textbf{Pre-processing} & \textbf{Method type} & \textbf{Data augmentation} & \textbf{External Dataset used} & \textbf{Post-processing} \\
\midrule
Baseline (T0) & IN & Image-based Two-stage & SA & No & Retain the largest connected components \\
\hline
Ganjie (T1) & Crop, resized to 224*224 & Video-based End-to-end & None & No & None \\
\hline
VCBiC (T2) & IN, resized to 256*256 & Image-based Two-stage & AutoAugmentation & PSFHS & Test-time augmentation, Ensembled predictions \\
\hline
BioMedia (T3) & IN & Image-based Two-stage & SA, CA, GN, GD, CJ & JNU-IFM, PSFHS & Retain the largest connected components \\
\hline
CQUT-Smart (T4) & Resized to 256*256 & Image-based Two-stage& SA, CA & ImageNet & Retain the largest connected components \\
\hline
nkdinsdale95 (T5) & None & Image-based Two-stage& SA, GN, GE, CA & ImageNet & MC, ED, pruning segmentation, Ellipse fitting \\
\hline
HHL\_hotpot (T6) & None & Image-based Two-stage& SA & None & Retain the largest connected components \\
\hline
Seilkbay (T7) & IN, Resized to 224*224 & Image-based Two-stage & SA & ImageNet & MC, ED, approximate mean squared ellipse \\
\bottomrule
\end{tabular}
\label{tab:Table5}
\end{table*}
\begin{table*}[!ht]
\centering
\caption{Network architectures and training details of the benchmarked algorithms. FCN: Fully Convolutional Networks; CE: Cross-Entropy loss; IoU: Intersection over Union loss; DC: Dice loss; SGD: Stochastic Gradient Descent; MSE: Mean Squared Error; DSSA: Dual Sparse Selection Attention; MFAM: Multi-Frequency Attention Module; EMC: an ensemble model of EfficientNetB0, ConvNeXt, ResNet-18, VGG-11 and DenseNet-196; EMS: an ensemble model of DeepLabv3+, UNet++ and MA-Net; VST: Video Swin Transformer; CV: cross-validation.}
\label{tab:Table6}

\fontsize{7}{8}\selectfont
\setlength{\tabcolsep}{3.2pt}
\renewcommand{\arraystretch}{1.05}

\begin{tabular}{@{}lcccccclll@{}}
\hline
\multirow{2}{*}{\textbf{Teams}} &
\multicolumn{3}{c}{\textbf{Architecture / Loss function / Input size}} &
\multirow{2}{*}{\textbf{Training}} &
\multirow{2}{*}{\textbf{Ensemble}} &
\multirow{2}{*}{\textbf{Optimizer}} &
\multirow{2}{*}{\textbf{Learning rate}} &
\multirow{2}{*}{\textbf{CV}} &
\multirow{2}{*}{\textbf{Device}} \\
\cline{2-4}
& \textbf{Classification} & \textbf{Segmentation} & \textbf{Biometry} &  &  &  &  &  &  \\
\hline

\multirow{3}{*}{Baseline (T0)} & FCN & 2D UNet & -- & \multirow{3}{*}{Two-stage} & \multirow{3}{*}{No} & \multirow{3}{*}{AdamW} & \multirow{3}{*}{$1\times10^{-4}$ (decay)} & \multirow{3}{*}{No} & \multirow{3}{*}{RTX 3090} \\
& CE & CE+DC & -- &  &  &  &  &  &  \\
& $512\times512$ & $512\times512$ & -- &  &  &  &  &  &  \\
\hline

\multirow{3}{*}{Ganjie (T1)} & VST & VST & VST & \multirow{3}{*}{One-stage} & \multirow{3}{*}{No} & \multirow{3}{*}{AdamW} & \multirow{3}{*}{$3\times10^{-5}$} & \multirow{3}{*}{No} & \multirow{3}{*}{Tesla A100} \\
& CE & CE+DC & MSE &  &  &  &  &  &  \\
& $224\times224$ & $224\times224$ & $224\times224$ &  &  &  &  &  &  \\
\hline

\multirow{3}{*}{ViCBiC (T2)} & ResNet & DeepLabv3 & -- & \multirow{3}{*}{Two-stage} & \multirow{3}{*}{Yes} & \multirow{3}{*}{AdamW} & \multirow{3}{*}{$1\times10^{-4}$} & \multirow{3}{*}{No} & \multirow{3}{*}{TITAN RTX} \\
& CE & CE+DC & -- &  &  &  &  &  &  \\
& $256\times256$ & $512\times512$ & -- &  &  &  &  &  &  \\
\hline

\multirow{3}{*}{BioMedIA (T3)} & FCN & 2D UNet & -- & \multirow{3}{*}{One-stage} & \multirow{3}{*}{No} & \multirow{3}{*}{Schedule-free} & \multirow{3}{*}{Dynamic} & \multirow{3}{*}{Yes} & \multirow{3}{*}{Tesla A100} \\
& CE & CE+DC & -- &  &  &  &  &  &  \\
& $512\times512$ & $512\times512$ & -- &  &  &  &  &  &  \\
\hline

\multirow{3}{*}{CQUT-Smart (T4)} & FCN & 2D UNet (DSSA) & -- & \multirow{3}{*}{Two-stage} & \multirow{3}{*}{No} & \multirow{3}{*}{Adam} & \multirow{3}{*}{$1\times10^{-4}$} & \multirow{3}{*}{Yes} & \multirow{3}{*}{RTX 3090} \\
& CE & CE+DC & -- &  &  &  &  &  &  \\
& $256\times256$ & $512\times512$ & -- &  &  &  &  &  &  \\
\hline

\multirow{3}{*}{nkdinsdale95 (T5)} & EMC & EMS & -- & \multirow{3}{*}{Two-stage} & \multirow{3}{*}{Yes} & \multirow{3}{*}{Adam} & \multirow{3}{*}{$1\times10^{-4}$ (decay)} & \multirow{3}{*}{Yes} & \multirow{3}{*}{Tesla A100} \\
& CE & CE+DC & -- &  &  &  &  &  &  \\
& $512\times512$ & $512\times512$ & -- &  &  &  &  &  &  \\
\hline

\multirow{3}{*}{HHL\_hotpot (T6)} & FCN & 2D UNet (MFAM) & -- & \multirow{3}{*}{Two-stage} & \multirow{3}{*}{No} & \multirow{3}{*}{AdamW} & \multirow{3}{*}{$1\times10^{-4}$ (decay)} & \multirow{3}{*}{No} & \multirow{3}{*}{RTX 3090} \\
& CE & CE+DC & -- &  &  &  &  &  &  \\
& $512\times512$ & $512\times512$ & -- &  &  &  &  &  &  \\
\hline

\multirow{3}{*}{Serikbay (T7)} & ResNet-50 & LinkNet & -- & \multirow{3}{*}{Two-stage} & \multirow{3}{*}{No} & \multirow{3}{*}{SGD} & \multirow{3}{*}{$1\times10^{-4}$} & \multirow{3}{*}{No} & \multirow{3}{*}{RTX 3090} \\
& CE & CE+DC & -- &  &  &  &  &  &  \\
& $224\times224$ & $512\times512$ & -- &  &  &  &  &  &  \\
\hline
\end{tabular}
\end{table*}

Within the IUGC, deep learning has garnered the most interest and has likewise demonstrated significant potential. Consistent with other challenges \citep{bai_2024_14328192,sappia2025acouslic,bai2025psfhs}, the keys to success involve the employment of preprocessing, suitable network architectures and loss functions, data augmentation, learning strategies, and post-processing. In this section, we examine the benchmarked methods based on these five aspects. \textbf{Table~\ref{tab:Table5}} and \textbf{Table~\ref{tab:Table6}} outline the key techniques employed by these methods.
\subsubsection{Proprocessing}
Preprocessing can decrease data complexity and enable models to learn target knowledge without being influenced by unnecessary variations. Widely employed techniques include image resizing and intensity normalization.

 To improve algorithm accuracy and reduce resource consumption, most teams adopt image-based processing strategies and/or adapt to the input requirements of pre-trained networks by reducing image resolution. For example, Team T1 uses video clips containing 4-frame images as input while setting the image size to 224×224 pixels, thus balancing computational resource usage and the ability to capture dynamic noise features in consecutive frames; Team T6 initializes the network with ImageNet pre-trained weights and trains the classification network using 224×224 pixel images \citep{deng2009imagenet}.

Intensity normalization seeks to convert the intensity ranges of images to a uniform scale. Z-score, a common and straightforward method, normalizes data to have a mean of zero and a standard deviation of one, adopted by T3 and T7; another approach involves linearly scaling an image's intensity range to [0,1], utilized by T2. More sophisticated preprocessing includes applying contrast enhancement to images. For instance, T1 used histogram equalization to boost contrast. For a summary of all teams, refer to \textbf{Table~\ref{tab:Table5}} for details.

\subsubsection{Architecture and loss function}
In this challenge, most participating teams adopted a multi-task model architecture based on the T0 framework, with the T0’s segmentation branch using a classic UNet architecture and the classification branch employing a traditional Fully Convolutional Network (FCN) \citep{shelhamer2016fully}—for example, Teams T3, T4, and T6 all used architectural designs similar to the baseline model. Some teams introduced innovative network combinations, such as Team T2 using DeepLabV3 for the segmentation branch and ResNet for the classification branch, and Team T7 employing LinkNet for segmentation and ResNet-50 for classification. Notably, Team T5 improved performance through a multi-model ensemble strategy, with its segmentation branch consisting of an ensemble model of EfficientNetB0 \citep{tan2019efficientnet}, ConvNeXt \citep{woo2023convnext}, ResNet-18, VGG-11, and DenseNet and its classification branch integrating DeepLabV3+ \citep{chen2017deeplab}, Unet++ \citep{zhou2019unet++}, and MA-Net \citep{han2024ma} architectures. Particularly, Team T1 adopted a unique unified design that not only integrated segmentation, classification, and ultrasound parameter measurement tasks into one framework but also achieved an end-to-end video processing solution via Video Swin Transformer \citep{liu2022video}, significantly differentiating it from other teams (\textbf{Table~\ref{tab:Table6}}).

The loss function functions as a key quantitative measure for assessing the consistency between a model's predictions and the actual target values, serving as the guiding criterion for the model's learning process and facilitating iterative enhancements in performance. In the IUGC, which involved three primary tasks—classification, segmentation, and measurement—all participating teams used the CE loss function for the classification task. For the segmentation task, in contrast, most teams, including T0, T2, T4, T5, T6, and T7, adopted a combined loss function integrating CE and DC, while Team T1 utilized a composite loss function with Intersection over Union (IoU) \citep{he2021alpha}, CE, and DC, and Team T3 relied solely on the Weighted Focal Dice loss function. Notably, in the measurement task, only Team T1 devised a specialized deep learning method, employing the Mean Squared Error (MSE) as the loss function (\textbf{Table~\ref{tab:Table6}}). Furthermore, while most teams opted for a two-stage training approach of segmentation followed by classification, Teams T1 and T3 implemented an end-to-end multi-task training strategy, which required the use of weighted composite loss functions, with the specific weight configurations thoroughly described in their respective publications \citep{gan2024accurate,alnasi2024multi}.
\subsubsection{Data Augmentation}
\begin{table*}[!ht]
\centering
\caption{Summary of the quantitative evaluation results of classification, segmentation and biometry tasks by the eight teams. Note that here DSC, ASSD, and HD represent the overall segmentation performance for the two targets, namely the pubic symphysis (PS) and the fetal head (FH). \textbf{The best results for each metric are highlighted in bold.}}
\label{tab:Table7}
\fontsize{7}{8} \selectfont
\begin{tabular}{@{}l|cccc|ccc|cc@{}}
\toprule 
\multirow{2}{*}{\textbf{Teams}} & \multicolumn{4}{c|}{\textbf{Classification}} & \multicolumn{3}{c|}{\textbf{Segmentation}} & \multicolumn{2}{c}{\textbf{Biometry}} \\
\cmidrule(r){2-5} \cmidrule(r){6-8} \cmidrule(l){9-10}
 & {\textbf{ACC}$\uparrow$} & {\textbf{F1}$\uparrow$} & {\textbf{AUC}$\uparrow$} & {\textbf{MCC}$\uparrow$} & {\textbf{DSC}$\uparrow$} & {\textbf{ASSD}$\downarrow$} & {\textbf{HD}$\downarrow$} & {\textbf{$\Delta$AoP}$\downarrow$} & {\textbf{$\Delta$HSD}$\downarrow$} \\
\midrule
\multirow{2}{*}{Baseline (T0)} 
& 0.4799 & 0.4515 & 0.4940 & 0.0758 & 0.7868 & 22.6453 & 89.0980 & 13.1965 & 19.7183 \\
& (0.2341) & (0.2163) & (0.3323) & (0.2893) & (0.1026) & (16.7296) & (63.9214) & (17.0165) & (15.2803) \\
\cline{1-10}

\multirow{2}{*}{Ganjie (T1)} 
& \textbf{0.7441} & \textbf{0.7555} & \textbf{0.7802} & \textbf{0.3648} & 0.8475 & 12.9975 & 38.7210 & 10.4283 & 11.4523 \\
& (0.2224) & (0.2278) & (0.3355) & (0.3553) & (0.0638) & (20.2825) & (31.6393) & (11.0271) & (9.4196) \\
\cline{1-10}

\multirow{2}{*}{ViCBiC (T2)} 
& 0.5670 & 0.6306 & 0.6868 & 0.1286 & \textbf{0.8857} & \textbf{9.4349} & \textbf{28.4152} & 9.4899 & \textbf{10.3878} \\
& (0.2453) & (0.2717) & (0.3363) & (0.2319) & (0.0476) & (8.5084) & (22.1607) & (9.7362) & (8.9781) \\
\cline{1-10}

\multirow{2}{*}{BioMedIA (T3)} 
& 0.6619 & 0.5422 & 0.7126 & 0.2145 & 0.8633 & 12.2203 & 41.1129 & \textbf{9.1557} & 11.6758 \\
& (0.2499) & (0.3740) & (0.3365) & (0.3200) & (0.0663) & (5.8450) & (19.8497) & (8.9648) & (18.6454) \\
\cline{1-10}

\multirow{2}{*}{CQUT-Smart (T4)} 
& 0.6319 & 0.6780 & 0.6225 & 0.2423 & 0.8535 & 11.0677 & 37.3815 & 10.7543 & 10.7867 \\
& (0.1900) & (0.2088) & (0.3028) & (0.2910) & (0.0596) & (5.1738) & (25.7057) & (10.0259) & (9.6300) \\
\cline{1-10}

\multirow{2}{*}{nkdinsdale95 (T5)} 
& 0.6789 & 0.7488 & 0.7571 & 0.2519 & 0.8169 & 14.3876 & 40.4391 & 15.3364 & 15.2268 \\
& (0.2386) & (0.2056) & (0.3355) & (0.3034) & (0.0906) & (15.7575) & (23.8167) & (16.4796) & (16.2635) \\
\cline{1-10}

\multirow{2}{*}{HHL\_hotpot (T6)} 
& 0.5707 & 0.6930 & 0.4383 & 0.0000 & 0.4767 & 82.3370 & 242.0712 & 62.4452 & 58.0625 \\
& (0.2515) & (0.2140) & (0.1644) & (0.0000) & (0.1217) & (42.5185) & (51.6231) & (31.8662) & (31.8337) \\
\cline{1-10}

\multirow{2}{*}{Serikbay (T7)} 
& 0.4293 & 0.0000 & 0.4410 & 0.0000 & 0.7263 & 56.2141 & 134.7939 & 40.1376 & 89.2053 \\
& (0.2515) & (0.0000) & (0.2499) & (0.0000) & (0.0866) & (11.7431) & (26.2063) & (29.4344) & (22.6194) \\
\bottomrule
\end{tabular}
\end{table*}

Data augmentation serves a vital function in ultrasound image processing by tackling the shortage of annotated clinical datasets—constrained by privacy issues and the time-intensive acquisition process—while also enhancing model generalization to mitigate overfitting through methods such as geometric transformations, intensity adjustments, and synthetic data generation \citep{2022A}. It simulates real-world ultrasound challenges such as speckle noise, shadowing, and probe movement-related variations, enhances performance for detecting rare anomalies or handling class imbalances, and reduces the need for costly data collection and annotation, thereby addressing ethical barriers. Additionally, it supports transfer learning and cross-domain adaptation between different ultrasound applications, proving vital for tasks like segmentation, detection, and classification by expanding training data and mimicking diverse imaging conditions. We categorize the augmentation techniques into two types, namely online and offline augmentation.

The online data augmentation techniques employed in this challenge encompass a wide array of operations, including translation, rotation, scaling, shearing, auto-contrast adjustment, inversion, equalization, solarization, posterization, contrast adjustment, color adjustment, brightness adjustment, sharpness enhancement, cutout, sample pairing, Gaussian noise addition, gamma enhancement, elastic deformation, grid distortion, and color jitter. While most teams such as T0, T6, and T7 resorted to simple data augmentation methods like rotation and flipping, Team T2 implemented nearly all available techniques through AutoAugmentation, where a sub-strategy—comprising two randomly selected operations (e.g., translation, rotation, shearing) with defined probabilities and intensities—was applied to each image in every mini-batch (\textbf{Table~\ref{tab:Table6}}). Notably, Team T3 conducted a systematic investigation into the impact of three augmentation scales (minimal, medium, and complex) on model performance, with results indicating that medium augmentation improved accuracy by approximately 15\% compared to minimal augmentation, and complex augmentation further enhanced accuracy by around 24\% relative to the medium scale \citep{alnasi2024multi}.

The offline augmentation primarily involves the utilization of external data or synthetic data generation, with most teams integrating external datasets: T2 and T3 leveraged the PSFHS challenge dataset \citep{chen2024psfhs,lu2022jnu}, while T4, T5, and T7 adopted pre-trained weights from ImageNet (\textbf{Table~\ref{tab:Table5}}). Notably, Team T3 stood out by employing synthetic data through a pseudo-labeling strategy with nnUNet: they trained nnUNet using 5-fold cross-validation on 2,575 labeled positive frames (each fold running for 1,000 epochs with the tool’s default 2D segmentation parameters), then ensembled the five resulting models by averaging their softmax outputs to generate pseudo-labels for unlabeled positive frames—after visually inspecting a random subset for plausibility, this process yielded pseudo-labels for 22,544 images \citep{alnasi2024multi}.

\subsubsection{Specification of the Learning Process}
As \textbf{Table~\ref{tab:Table6}} shows, only T1 adopted an end-to-end approach, while other teams implemented their works in a two-stage manner: first, identifying the standard plane and segmenting FH and PS; second, calculating ultrasound parameters based on the segmentation results. Furthermore, the approaches used to identify the standard plane and segment the target region differed, which could account for the discrepancies in their results. For example, T3 and T1 adopted an end-to-end training approach for multiple tasks, whereas T0, T2, T4, T5, T6, and T7 first trained the segmentation branch of the framework and then trained the classification branch. T5 used an ensemble of multiple models for both classification and segmentation, while other teams employed only a single model. Among them, T2 conducted a comparative analysis of the performance of different models and found that ResNet, SqueezeNet, and PoolFormer performed well in classification tasks, while UNet++, PANNet, LinkNet, and DeepLabV3 showed better results.

Due to the scarcity of training data in the IUGC, predefining the optimal model is challenging. Furthermore, models trained on different samples or images from various views may acquire distinct knowledge. Therefore, several ensemble learning strategies were employed in this challenge to mitigate model uncertainty and enhance generalization capabilities. For example, T5 adopted an ensemble model composed of three deep learning networks to complete classification and segmentation tasks, respectively. They compared various combinations including VGG-11, VGG-16, ResNet-18, ResNet-152, WideResNet-50, DenseNet-196, EfficientNet, EfficientNet-V2, and ConvNeXT, and ultimately found that an ensemble of EfficientNetB0, ConvNeXt, ResNet-18, and VGG-11 achieved the best classification results \citep{ramesh2024intrapartum}. They also compared combinations including UNet, UNet++, DeepLab V3+, and MA-Net, and determined that an ensemble of DeepLabV3+, UNet++, and MA-Net yielded the optimal segmentation performance. Unlike T5's multi-model integration approach, T2 amplified data quantity by horizontally and vertically flipping input images and averaged multiple predictions during inference \citep{tan2024classification}.
\subsubsection{Post-processing}
The post-processing strategies adopted by various teams differ in complexity and intent, reflecting their specific segmentation goals and confidence in model outputs. Teams like T0, T3, T4, and T6 retain the largest connected components to eliminate spurious small regions and isolate primary anatomical structures, ensuring cleaner segmentation for contiguous objects. In contrast, T5 and T7 employ more sophisticated pipelines involving morphological closing to fill holes, edge-detection to refine boundaries, iterative pruning to remove inconsistent segments, and ellipse approximation to fit geometric shapes for precise parameter measurements. T2 utilizes a data-driven approach with test-time augmentation, applying multiple transformations to test images and aggregating predictions to reduce variance, complemented by quantity amplification and ensemble methods to enhance robustness. Notably, T1 employs an end-to-end framework without post-processing, a design choice that preserves the model’s direct output while leveraging the single-stage optimization capability of end-to-end pipelines to eliminate the need for separate refinements (\textbf{Table~\ref{tab:Table6}}). These diverse approaches highlight the trade-off between simplicity/computational efficiency and precision-driven complex refinements, with practices like connected component analysis underscoring their role as fundamental cleanup techniques.

\section{Results}
\begin{figure}[htbp]
	\centering
\includegraphics[width=0.4\textwidth]{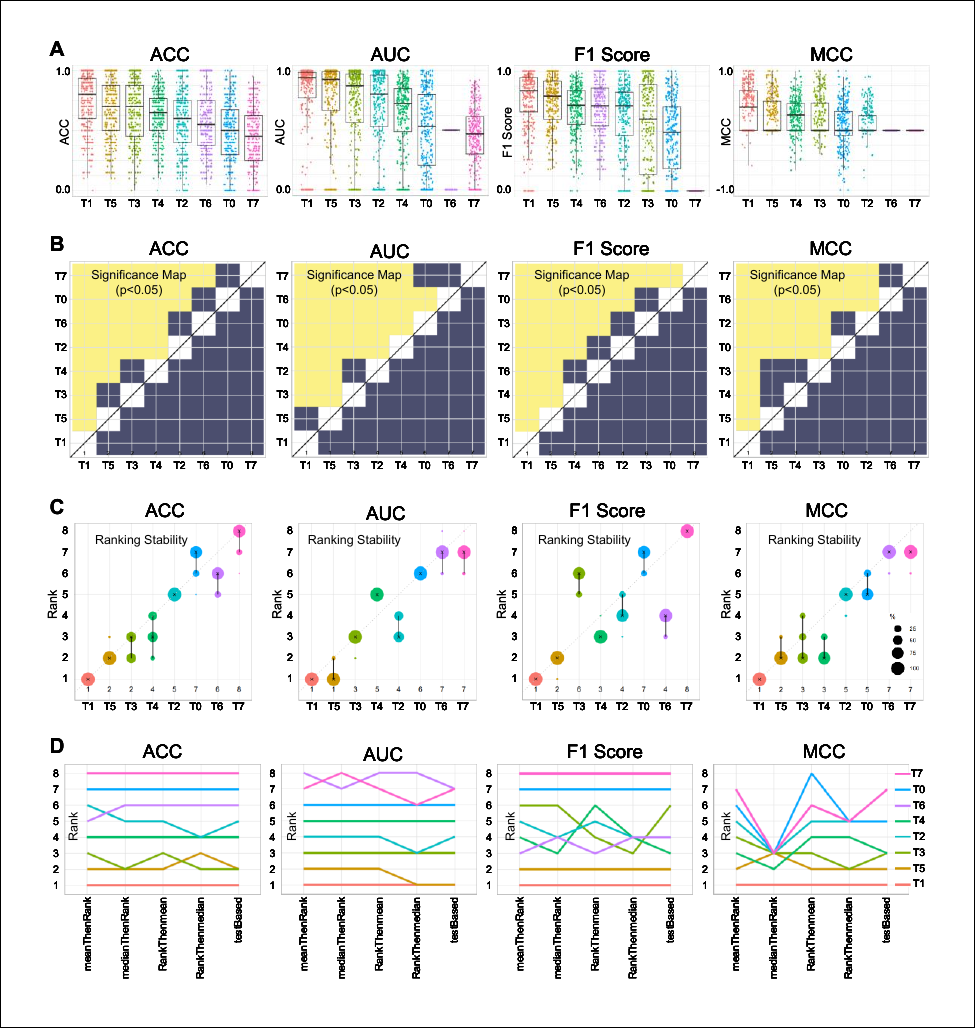}
\caption{Evaluation results of eight teams' methods in the classification task based on ACC (first column), AUC (second column), F1 Score (third column), and MCC (fourth column). (A) Dot plots and boxplots for visualizing the evaluation metric data separately for each algorithm. (B) Blob plots for visualizing ranking stability
based on bootstrap sampling. (C) Significance maps for visualizing the results of significance testing. (D) Line plots for
visualizing rankings robustness across different ranking methods. See Section 3.3.2 for details.}
\label{Figure5}
\end{figure}
In this section, we summarize the performance of the eight methods and their ranking results. For the performance evaluation in Section 5.1, we use nine metrics to assess the performance of the eight methods in the three tasks. For each evaluation metric, we compare the performance of each method from four aspects: distribution of evaluation metric data, significance maps, ranking stability and ranking robustness. For the ranking analysis in Section 5.2, we respectively analyze the rankings of the eight methods in each task and their comprehensive rankings in the entire challenge.
\subsection{Overall Performance}
Here, we synthesize the performance of the eight methods across the tasks of classification, segmentation, and biometry. Classification performance is assessed via ACC, AUC, F1 score, and MCC; segmentation performance is measured by DSC, ASSD, and HD; and biometry performance is evaluated through $\Delta$AoP and $\Delta$HSD. The quantitative results of the eight methods across the nine indicators are shown in \textbf{Table~\ref{tab:Table7}}. Teams Ganjie (T1), ViCBiC (T2), and BioMedIA (T3) demonstrate excellent performance in the classification, segmentation, and measurement tasks, respectively.

\subsubsection{Performance on the Classification Task}
\begin{figure}[htbp]
	\centering
\includegraphics[width=0.4\textwidth]{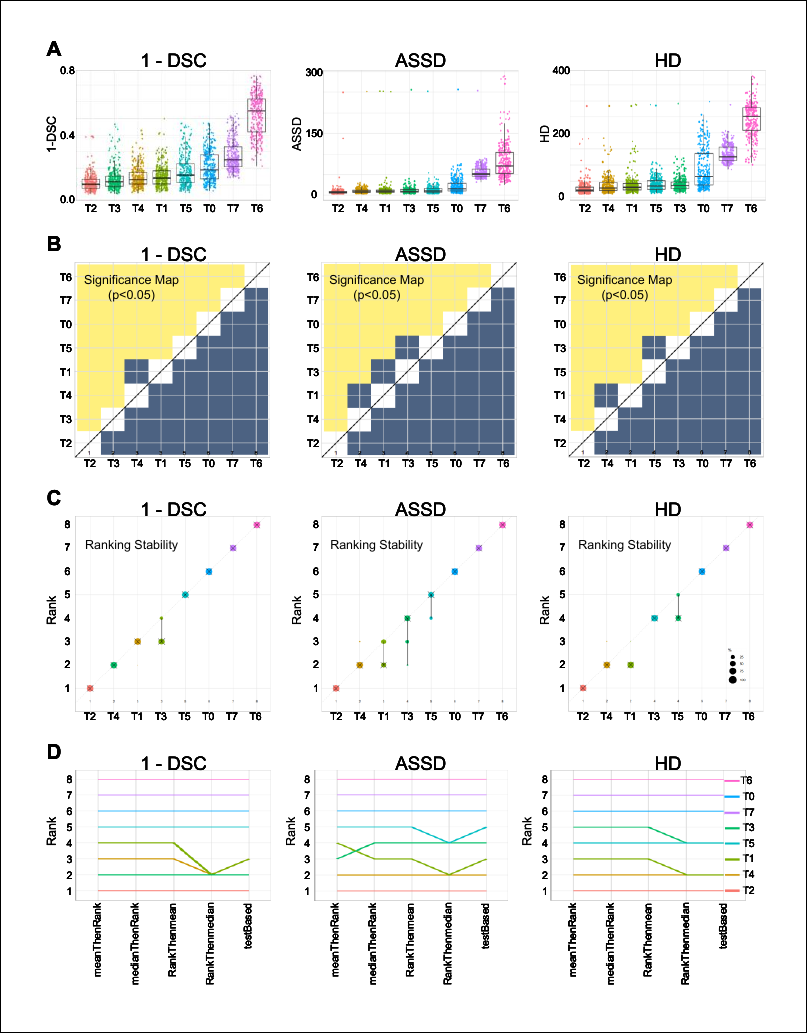}
\caption{Evaluation results of eight teams' methods in the segmentation task based on DSC (first column), ASSD (second column), and HD (third column). (A) Dot plots and boxplots for visualizing the evaluation metric data separately for each algorithm. (B) Blob plots for visualizing ranking stability
based on bootstrap sampling. (C) Significance maps for visualizing the results of significance testing. (D) Line plots for
visualizing rankings robustness across different ranking methods. See Section 3.3.2 for details.}
\label{Figure6}
\end{figure}
\textbf{Fig.~\ref{Figure5}} illustrates the classification performance of eight algorithms on the test set with ACC, AUC, F1 score, and MCC. For each metric, we first utilize dot plots and boxplots to visualize the distribution of evaluation metric data for each individual algorithm (\textbf{Fig.~\ref{Figure5}A}). Subsequently, significance maps are employed to present the results of significance testing, enabling a clear comparison of the statistical significance among algorithms (\textbf{Fig.~\ref{Figure5}B}). Blob plots are subsequently employed to illustrate the stability of rankings derived from bootstrap sampling, offering insights into how consistently algorithm rankings perform across various resampling situations (\textbf{Fig.~\ref{Figure5}C}). Finally, line plots are adopted to demonstrate the ranking robustness of each algorithm across various ranking methods, highlighting how stable the rankings remain when different analytical approaches are applied (\textbf{Fig.~\ref{Figure5}D}).

T1 outperforms other methods in the classification task, with ACC, F1 score, AUC, and MCC values of 0.7441, 0.7555, 0.7802, and 0.3648, respectively (\textbf{Table~\ref{tab:Table7}} and \textbf{Fig.~\ref{Figure5}A}). Statistical significance analysis reveals that T1 significantly outperforms other algorithms in ACC, F1 score, and MCC (yellow, p $<$ 0.05), while demonstrating comparable performance in AUC (blue) (\textbf{Fig.~\ref{Figure5}B}). In terms of ranking stability, T1 ranks first across all four metrics (\textbf{Fig.~\ref{Figure5}C}). Additionally, when comparing rankings based on different schemes, T1 consistently maintains the top position, whereas most other algorithms exhibit fluctuations in their rankings (\textbf{Fig.~\ref{Figure5}D}).

\subsubsection{Performance on the Segmentation Task}
\begin{figure}[htbp]
	\centering
\includegraphics[width=0.35\textwidth]{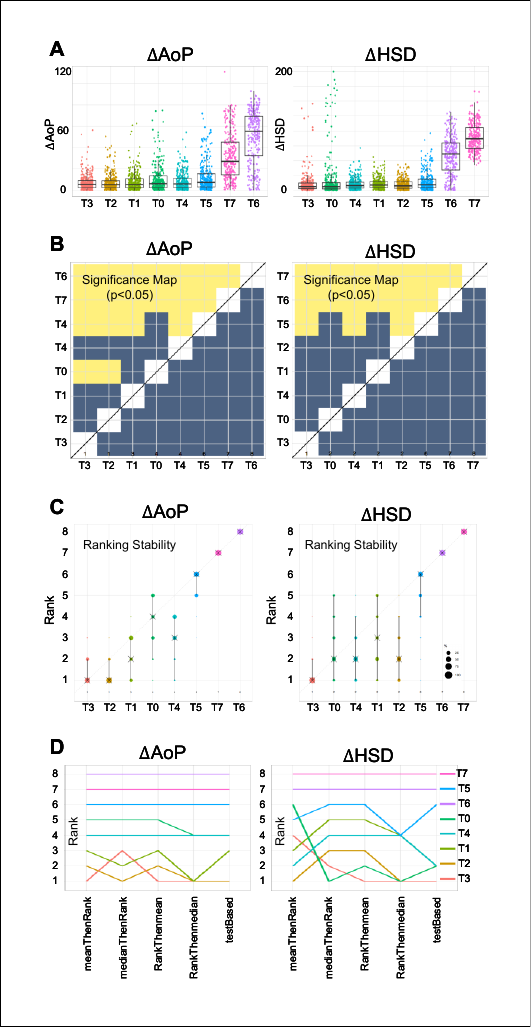}
\caption{Evaluation results of eight teams' methods in the biometry task based on $\Delta$AoP (first column), and $\Delta$HSD (second column). (A) Dot plots and boxplots for visualizing the evaluation metric data separately for each algorithm. (B) Blob plots for visualizing ranking stability
based on bootstrap sampling. (C) Significance maps for visualizing the results of significance testing. (D) Line plots for
visualizing rankings robustness across different ranking methods. See Section 3.3.2 for details.}
\label{Figure7}
\end{figure}
\textbf{Fig.~\ref{Figure6}} illustrates the segmentation performance of eight algorithms on the test set. T2 outperforms other methods, with DSC, ASSD, and HD values of 0.8857,	9.4349 and 28.4152, respectively (\textbf{Table~\ref{tab:Table8}} and \textbf{Fig.~\ref{Figure6}A}). Statistical significance analysis reveals that T1 significantly outperforms other algorithms in 1-DSC, ASSD, and HD (yellow, p $<$ 0.05)(\textbf{Fig.~\ref{Figure6}B}). In terms of ranking stability, T1 ranks first across all three metrics (\textbf{Fig.~\ref{Figure6}C}). Additionally, when comparing rankings based on different schemes—including MedianThenRank, MeanThenRank, RankThenMedian, RankThenMean, and testBased—T1 consistently maintains the top position, whereas some algorithms exhibit fluctuations in their rankings (\textbf{Fig.~\ref{Figure6}D}).
\subsubsection{Performance on the Biometry Task}
\begin{figure}[htbp]
	\centering
\includegraphics[width=0.35\textwidth]{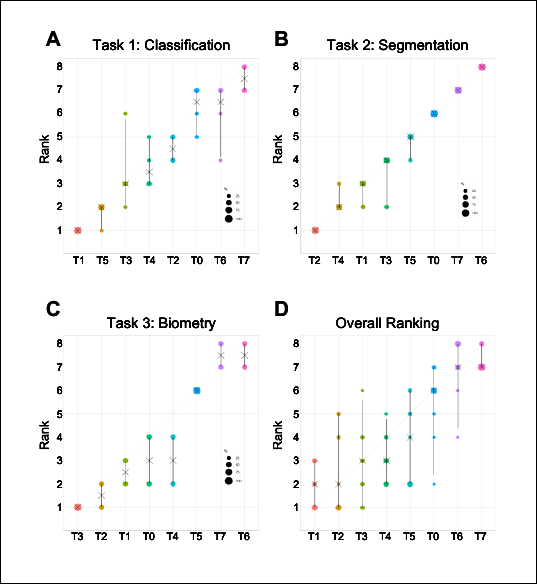}
\caption{Ranking Results. (A) Ranking results in the classification task; (B) Ranking results in the segmentation task; (C) Ranking results in the biometry task; (D) Comprehensive ranking results. See Section 3.3.2 for details.}
\label{Figure8}
\end{figure}
\textbf{Fig.~\ref{Figure7}} illustrates the biometry performance of eight algorithms on the test set. The Biometry performance is evaluated using two metrics: $\Delta$AoP and $\Delta$HSD. Team T3 achieves the smallest \(\Delta\)AoP (9.1557), whereas Team T2 achieves the smallest \(\Delta\)HSD (10.3878) (\textbf{Fig.~\ref{Figure7}A}). For \(\Delta\)AoP, statistical significance analysis reveals that T3 significantly outperforms T0, T5, T6, and T7, while demonstrating comparable performance to T1, T2, and T4. For \(\Delta\)HSD, the analysis shows T3 significantly outperforms T5, T6, and T7, with comparable results to T0, T1, T2, and T4 (\textbf{Fig.~\ref{Figure7}B}). In terms of ranking stability, while T3's rankings tend to cluster around the top position, its ranking stability is low (\textbf{Fig.~\ref{Figure7}C}). Additionally, when comparing rankings across different schemes, T3 ranks first in all \(\Delta\)AoP evaluations except for a third-place ranking in MedianThenRank. For \(\Delta\)HSD, T3 ranks first in most schemes, with exceptions including fourth place in MeanThenRank and second place in MedianThenRank. Notably, T2 demonstrates strong performance in specific scenarios: it ranks first in \(\Delta\)AoP under MedianThenRank, RankThenMedian, and testBased, and in \(\Delta\)HSD under MeanThenRank, RankThenMedian, and testBased (\textbf{Fig.~\ref{Figure7}D}).
\begin{table*}[ht]
\centering
\caption{Inter-annotator agreement analysis. From the test dataset, 50 samples were randomly selected from each of the three sources (i.e., JNU, SYSU and SMU). Three annotators participated in the labeling process, with the first annotator's results serving as the ground truth. The annotations of annotators 2 and 3 were evaluated against this reference, and the intraclass correlation coefficient (ICC) was calculated to measure inter-rater reliability.}
\label{tab:Table8}
\fontsize{7}{8} \selectfont
\begin{tabular}{@{}llcccc@{}}
\toprule
\textbf{Tasks}                           & \textbf{ICC}            & \textbf{JNU (n=50)} & \textbf{SYSU (n=50)} & \textbf{SMU (n=50)} & \textbf{All (n=150)} \\
\midrule
\multirow{4}{*}{Classification} & ICC (ACC)      & 0.99       & 0.96        & 0.85       & 0.92        \\
                                & ICC (AUC)      & 0.99       & 0.96        & 0.95       & 0.96        \\
                                & ICC (F1)       & 0.99       & 0.98        & 0.86       & 0.94        \\
                                & ICC (MCC)      & 0.99       & 0.96        & 0.77       & 0.89        \\
\hline
\multirow{6}{*}{Segmentation}   & ICC (DSC\_PS)  & 0.84       & 0.71        & 0.84       & 0.82        \\
                                & ICC (DSC\_FH)  & 0.64       & 0.59        & 0.50       & 0.57        \\
                                & ICC (ASSD\_PS) & 0.88       & 0.74        & 0.83       & 0.86        \\
                                & ICC (ASSD\_FH) & 0.58       & 0.58        & 0.46       & 0.47        \\
                                & ICC (HD\_PS)   & 0.94       & 0.75        & 0.89       & 0.91        \\
                                & ICC (HD\_FH)   & 0.62       & 0.50        & 0.28       & 0.47        \\
\hline
\multirow{2}{*}{Biometry}       & ICC ($\Delta$AoP)     & 0.49       & 0.54        & 0.59       & 0.53        \\
                                & ICC ($\Delta$HSD)     & 0.48       & 0.38        & 0.48       & 0.46        \\
\bottomrule
\end{tabular}
\end{table*}
\begin{table*}[ht]
\centering
\caption{Results of the baseline method and the top-ranked method across three distinct source datasets. \textbf{The best results for each metric are highlighted in bold.}}
\label{tab:Table9}
\fontsize{7}{8} \selectfont
\begin{tabular}{@{}l|l|ccc|cccccc|cc@{}}
\toprule
\multirow{2}{*}{\textbf{Teams}} & \multirow{2}{*}{\textbf{Sources}} & \multicolumn{3}{c|}{\textbf{Classification}} & \multicolumn{6}{c|}{\textbf{Segmentation}} & \multicolumn{2}{c}{\textbf{Biometry}} \\
\cmidrule(r){3-5} \cmidrule(r){6-11} \cmidrule(l){12-13}
 &  & {\textbf{ACC}$\uparrow$} & {\textbf{AUC}$\uparrow$} & {\textbf{F1}$\uparrow$} & {\textbf{DSC\_PS}$\uparrow$} & {\textbf{DSC\_FH}$\uparrow$} & {\textbf{ASSD\_PS}$\downarrow$} & {\textbf{ASSD\_FH}$\downarrow$} & {\textbf{HD\_PS}$\downarrow$} & {\textbf{HD\_FH}$\downarrow$} & {\textbf{$\Delta$AoP}$\downarrow$} & {\textbf{$\Delta$HSD}$\downarrow$} \\
\midrule
\multirow{6}{*}{Baseline (T0)} 
& \multirow{2}{*}{JNU} 
& 0.6836 & 0.8664 & 0.7208 & 0.6939 & 0.9067 & 26.1792 & 10.2792 & 133.8031 & 31.8094 & 10.2190 & 11.8456 \\
&  & (0.2077) & (0.1814) & (0.2266) & (0.1907) & (0.0419) & (25.1389) & (4.3180) & (130.3072) & (17.1361) & (10.5417) & (10.5297) \\
\cline{2-13}

& \multirow{2}{*}{SYSU} 
& 0.6797 & 0.8679 & 0.7188 & 0.7741 & 0.8794 & 16.9782 & 12.2988 & 91.8806 & 40.3748 & 9.5574 & 9.1950 \\
&  & (0.1877) & (0.1780) & (0.2009) & (0.0693) & (0.0872) & (22.0109) & (7.0011) & (115.9075) & (22.5339) & (7.5543) & (7.1125) \\
\cline{2-13}

& \multirow{2}{*}{SMU} 
& 0.6880 & 0.5087 & \textbf{0.7628} & 0.5944 & 0.7921 & 56.1035 & 23.8519 & 238.5713 & 79.1492 & 19.6740 & 26.1078 \\
&  & (0.2911) & (0.4158) & (0.2412) & (0.2011) & (0.1491) & (45.9167) & (11.2215) & (152.1613) & (34.2716) & (17.1236) & (23.5715) \\
\hline

\multirow{6}{*}{Ganjie (T1)} 
& \multirow{2}{*}{JNU} 
& \textbf{0.7657} & \textbf{0.9168} & \textbf{0.7709} & \textbf{0.7464} & \textbf{0.9170} & \textbf{9.4456} & \textbf{9.2929} & \textbf{24.6326} & \textbf{28.4345} & \textbf{8.2284} & \textbf{9.6175} \\
&  & (0.1746) & (0.1520) & (0.2183) & (0.1475) & (0.0493) & (5.8381) & (5.4876) & (13.2775) & (17.8387) & (7.2571) & (9.9385) \\
\cline{2-13}

& \multirow{2}{*}{SYSU} 
& \textbf{0.7760} & \textbf{0.9205} & \textbf{0.7574} & \textbf{0.8106} & \textbf{0.9144} & \textbf{6.7862} & \textbf{9.3979} & \textbf{21.0951} & \textbf{29.4822} & \textbf{7.1727} & \textbf{9.0411} \\
&  & (0.1872) & (0.1591) & (0.2680) & (0.0813) & (0.0329) & (2.9593) & (3.6633) & (9.7958) & (13.2755) & (6.1284) & (6.1019) \\
\cline{2-13}

& \multirow{2}{*}{SMU}  
& \textbf{0.7120} & \textbf{0.5855} & 0.7414 & \textbf{0.6034} & \textbf{0.9042} & \textbf{46.4052} & \textbf{13.4713} & \textbf{91.7589} & \textbf{44.2756} & \textbf{18.3155} & \textbf{15.6353} \\
&  & (0.2752) & (0.4342) & (0.2779) & (0.2095) & (0.0445) & (139.2619) & (5.8848) & (154.5262) & (22.0674) & (17.1390) & (11.7940) \\
\bottomrule
\end{tabular}
\end{table*}

\subsection{Ranking Results}
In the IUGC 2024 challenge, we investigated not only rankings based on individual evaluation metrics but also task-based rankings and challenge-oriented rankings. Specifically, the classification task rankings were based on ACC, AUC, F1 score, and MCC; the segmentation task rankings relied on DSC, ASSD, and HD; the measurement task rankings were determined by  \(\Delta\)AoP and  \(\Delta\)HSD; and the comprehensive challenge ranking was a composite of the results from all three tasks.

\textbf{Fig.~\ref{Figure8}} presents the task-specific rankings and overall challenge ranking of each team. Team T1 tops the classification task (\textbf{Fig.~\ref{Figure8}A}), T2 leads the segmentation task (\textbf{Fig.~\ref{Figure8}B}), and T3 ranks first in the biometry task (\textbf{Fig.~\ref{Figure8}C}). In the comprehensive ranking, T1, T2, and T3 secure the top three positions, though their rankings exhibit instability: T1 fluctuates between 1st and 3rd place, T2 ranges from 1st to 5th, and T3 varies between 1st and 6th (\textbf{Fig.~\ref{Figure8}D}). Notably, T1 demonstrates more stable rankings than T2 and T3, maintaining the overall first position.
\section{Discussion}
This section discusses the IUGC results from six dimensions: annotation consistency, model performance on data from different sources, model training methods, data preprocessing approaches, network architectures, post-processing techniques, and research limitations.

\subsection{Inter-annotator Agreement Analysis}
To investigate the variability among different annotators, we randomly selected 150 samples from the test dataset: 50 from JNU, 50 from SYSU, and 50 from SMU. Three annotators were invited to independently annotate these samples. The annotations from the first annotator (serving as the ground truth for this challenge) were used as the reference standard to evaluate the results of the second and third annotators. The intraclass correlation coefficient (ICC) was employed to quantify inter-rater reliability.

\textbf{Table~\ref{tab:Table8}} presents the ICC values for JNU, SYSU, SMU, and the combined dataset All (JNU+SYSU+SMU), following standard interpretation criteria: ICC $<$ 0.2 denotes poor agreement, 0.2--0.4 indicates fair agreement, 0.4--0.6 represents moderate agreement, 0.6--0.8 signifies strong agreement, and 0.8--1.0 corresponds to very strong agreement. For the combined dataset, ICC values across all evaluation metrics exceeded 0.4 (moderate agreement), with classification metrics demonstrating particularly high consistency (ICC $>$ 0.8). In segmentation, PS structure annotations showed consistently higher agreement than FH annotations, including ICC (DSC\_PS) = 0.82 vs. ICC (DSC\_FH) = 0.57, ICC (ASSD\_PS) = 0.86 vs. ICC (ASSD\_FH) = 0.47, and ICC (HD\_PS) = 0.86 vs. ICC (HD\_FH) = 0.47. For ultrasound biometry, $\Delta$AoP ($\text{ICC}=0.53$) exhibited slightly higher reliability than $\Delta$HSD ($\text{ICC}=0.46$).

When analyzed by center, the JNU subset demonstrated strong agreement for most classification and segmentation metrics (ICC approaching or exceeding 0.6), while ultrasound parameters achieved moderate agreement (ICC $>$ 0.48). In SYSU, $\Delta$HSD showed fair agreement (ICC $<$ 0.4), whereas other metrics maintained moderate agreement (ICC $>$ 0.4), with PS segmentation again outperforming FH segmentation (ICC $>$ 0.7 vs. 0.4--0.6). In the SMU subset, HD\_FH exhibited fair agreement ($\text{ICC}=0.28$), while other metrics remained within the moderate range, and PS segmentation achieved consistently strong agreement (ICC $>$ 0.8), substantially exceeding FH-related metrics (0.28--0.50).

Overall, classification and PS segmentation annotations demonstrated strong inter-observer consistency (ICC $>$ 0.7), whereas FH segmentation and ultrasound biometry measurements exhibited moderate agreement (ICC = 0.4--0.6). This level of agreement reflects the inherent clinical difficulty of intrapartum ultrasound measurements, which are affected by dynamic fetal motion, probe-induced tissue deformation, acoustic shadowing, partial landmark visibility, and operator-dependent acquisition variability. Consequently, the observed ICC values primarily indicate intrinsic task ambiguity rather than deficiencies in the annotation protocol.

Although multi-expert consensus labeling represents an ideal annotation strategy, the current protocol follows standardized clinical measurement guidelines and was performed by experienced sonographers with independent double annotation and quality control procedures, which is consistent with common practice in prior intrapartum ultrasound benchmark studies. Therefore, while annotation uncertainty exists, the current reference standard remains clinically meaningful and suitable for comparative algorithm evaluation.

An important implication of this variability is the interpretation of benchmark ranking when algorithmic differences are on the order of 1--2 degrees. In this benchmark, all methods are evaluated under a unified annotation protocol and identical reference standards, ensuring fairness at the comparative level. Furthermore, our bootstrap-based ranking stability analysis demonstrates that the relative ordering of top-performing methods remains consistent across resampled test subsets despite annotation noise, indicating that the leaderboard remains robust for comparative evaluation even when absolute measurement uncertainty exists.

It should also be emphasized that inter-observer ICC does not constitute a strict upper bound on achievable algorithmic performance. While ICC quantifies pairwise human agreement, learning-based models trained on large-scale datasets capture population-level consensus patterns and can reduce random inter-observer variability through statistical aggregation. Nevertheless, annotation uncertainty imposes a practical performance ceiling, beyond which further numerical improvements may not directly translate into clinically meaningful gains.

These findings motivate future benchmark extensions, including multi-expert consensus labeling, probabilistic annotation fusion strategies, and uncertainty-aware evaluation protocols, which may further enhance annotation reliability, improve interpretability of performance differences, and better align benchmark evaluation with real-world clinical decision-making.

\subsection{Source-stratified Robustness and Ranking Stability}
\begin{figure}[htbp]
	\centering
\includegraphics[width=0.48\textwidth]{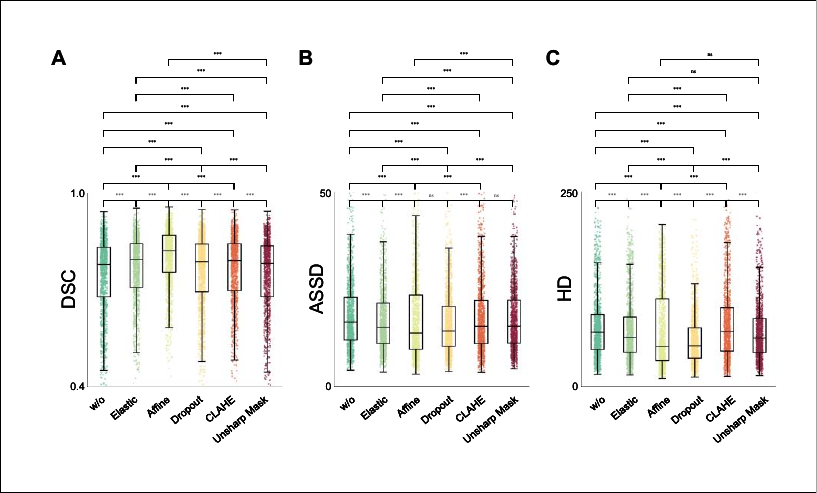}
\caption{Effects of Different Data Augmentation Methods on Segmentation Performance Based on DSC (A), ASSD (B), and HD (C). w/o - without data augmentation; Elastic - random spatial distortions applied to images; Affine - affine transformation (includes scaling, translation, shear, and rotation of images); Dropout -random selection of a rectangular area with pixels set to 0; Unsharp Mask - image contrast enhancement via unsharp masking; CLAHE - contrast-limited adaptive histogram equalization.}
\label{Figure9}
\end{figure}
\begin{table*}[ht]
\centering
\caption{Impact of different training frameworks on baseline model performance. The training architectures include ``segment-then-classify'' (training the segmentation model first followed by the classification model, C\_Then\_S) and ``classify-then-segment'' (training the classification model first followed by the segmentation model, S\_Then\_C). Note that here DSC, ASSD, and HD represent the overall segmentation performance for the two targets, namely the pubic symphysis (PS) and the fetal head (FH). \textbf{The best results for each metric are highlighted in bold.}}
\label{tab:Table10}
\fontsize{7}{8} \selectfont
\begin{tabular}{@{}lccccccc@{}}
\toprule
\textbf{Framework} & \textbf{ACC}$\uparrow$ & \textbf{AUC}$\uparrow$ & \textbf{F1}$\uparrow$ & \textbf{MCC}$\uparrow$ & \textbf{DSC}$\uparrow$ & \textbf{ASSD}$\downarrow$ & \textbf{HD}$\downarrow$ \\
\midrule
\multirow{2}{*}{C\_Then\_S} 
& 0.6579 
& 0.7048 
& 0.5359 
& \textbf{0.2467} 
& 0.7529 
& \textbf{20.2196} 
& 74.8858 \\
& (0.2483) 
& (0.3534) 
& (\underline{0.3757})*** 
& (0.3328) 
& (\underline{0.1089})*** 
& (13.5374) 
& (\underline{45.2165})*** \\
\hline
\multirow{2}{*}{S\_Then\_C} 
& \textbf{0.6718} 
& \textbf{0.7180} 
& \textbf{0.7290} 
& 0.2309 
& \textbf{0.8082} 
& 84.1366 
& \textbf{21.2971} \\
& (0.2388) 
& (0.3381) 
& (0.2260) 
& (0.3002) 
& (0.1018) 
& (56.8085) 
& (19.3117) \\
\bottomrule
\end{tabular}
\end{table*}

To analyze method performance across diverse data sources, we compared the baseline (T0) and winning (T1) methods on the three test subsets (JNU, SYSU, and SMU). In the classification and biometry tasks, T1 consistently outperformed T0 across all centers. For example, on JNU, T1 achieved an ACC of 0.7657 (vs. 0.6833 for T0), an F1-score of 0.7709 (vs. 0.7208), and a $\Delta$AoP of 8.2284 (vs. 10.2900), demonstrating superior feature discrimination and measurement precision. In contrast, segmentation performance of T1 and T0 was comparable on JNU, with DSC$_\mathrm{PS}$ values of 0.7465 and 0.7533, respectively, and similar HD$_\mathrm{FH}$ values (28.4345 vs. 29.6930), indicating comparable anatomical delineation capability (Table~\ref{tab:Table9}).

A pronounced performance gap is observed on the SMU (Esaote MyLab) dataset across all evaluated methods. While T1 exhibits stable performance on JNU and SYSU, its performance consistently degrades on SMU. Specifically, classification metrics drop to ACC = 0.7120, F1 = 0.7414, and AUC = 0.5851, while biometry errors increase substantially ($\Delta$AoP = 17.3890 and $\Delta$HSD = 15.6353). Segmentation performance also deteriorates markedly, with DSC$_\mathrm{PS}$ decreasing from 0.746 (JNU) and 0.811 (SYSU) to 0.603 on SMU, and HD$_\mathrm{FH}$ increasing to 43.2755. Importantly, this degradation pattern is consistently observed across all participating methods, indicating that SMU images are intrinsically more challenging rather than merely distributionally different.

Several factors may contribute to this systematic performance drop. First, SMU data were acquired using the Esaote MyLab system, which exhibits device-specific imaging characteristics, including differences in spatial resolution, beamforming behavior, and speckle noise statistics compared to the systems used at JNU and SYSU. Second, acquisition protocols and probe positioning strategies may vary across centers, affecting anatomical visibility and introducing heterogeneous shadowing artifacts. Third, differences in operator experience and scanning workflows may influence frame selection quality and image stability. Finally, patient-related factors such as maternal body habitus and pelvic anatomy distribution may further impact acoustic penetration and boundary visibility. Together, these factors likely compound the difficulty of segmentation and landmark localization on SMU data.

Given the limited number of acquisition devices and the imbalanced source composition, we further performed a source-stratified evaluation to examine ranking robustness under domain shift. For each center ($n=50$), we report task-wise metrics with mean$\pm$std and compute task-specific AvgRank and Overall AvgRank by converting all 15 evaluation metrics into ordinal ranks with consistent optimization directions. As summarized in \textbf{Supplementary Table~S1}, top-performing methods maintain consistently low Overall AvgRank across all sources, indicating stable leaderboard structure. In contrast, mid-tier and baseline methods exhibit substantially larger rank fluctuations, particularly on the most challenging source (SMU), reflecting increased sensitivity to device heterogeneity and domain shift. \textbf{Supplementary Fig. S1} further quantifies this effect, showing strong cross-source correlation for Overall AvgRank (Spearman's $\rho = 0.81$--$0.94$), confirming that global ranking remains largely preserved.

However, task-level analysis reveals notable differences. Segmentation shows the highest cross-source stability ($\rho > 0.93$), classification exhibits moderate stability ($\rho \approx 0.62$--$0.71$), while biometry demonstrates the lowest stability ($\rho \approx 0.58$--$0.63$), consistent with the larger AvgRank dispersion observed in \textbf{Supplementary Table S1}. These results indicate that source imbalance and device heterogeneity primarily affect high-level semantic recognition and downstream clinical measurement tasks, whereas pixel-level segmentation remains comparatively robust.

From a benchmarking perspective, these findings highlight the necessity of device-aware and center-aware evaluation protocols and demonstrate that current state-of-the-art intrapartum ultrasound algorithms remain sensitive to real-world clinical heterogeneity. Importantly, the pronounced SMU performance gap also underscores the value of including challenging acquisition domains in benchmark design, preventing overly optimistic performance estimates derived from homogeneous data and encouraging future research on domain generalization, device-adaptive learning, and multi-source harmonization strategies.

\subsection{Comparison of Different Training Approaches}
The baseline (T0) method adopts a model with a shared encoder and two task-specific decoder branches, corresponding to the segmentation and classification outputs, respectively. During training, the segmentation task is performed first to guide the encoder in learning structural and spatial features. The encoder is then frozen, and the classification branch is trained independently. In this setting, the classification task relies on the feature representations learned during segmentation \citep{zhou2024baseline}.

This design is partly inspired by clinical annotation practices. When annotating whether an image represents a standard plane, clinicians implicitly localize and outline FH and PS structures. From a technical perspective, although all training images are standard planes, we hypothesize that the segmentation task enables the encoder to develop discriminative representations that can generalize to unseen, non-standard images. Specifically, due to the structural differences, the encoder’s response to a non-standard plane may differ from that to a standard plane, and this difference becomes a critical cue for classification (\textbf{Table~\ref{tab:Table6}}).

To validate this hypothesis, we compared two training paradigms, both using a shared encoder and dual-decoder architecture. In the first paradigm, segmentation is trained first, followed by classification with the encoder frozen (i.e., Segmentation-Then-Classification, S\_Then\_C). In the second, the order is reversed: classification is trained first, then segmentation (i.e., Classification-Then-Segmentation, C\_Then\_S). This comparison allows us to examine how task ordering influences the learned representations and overall performance.

As shown in \textbf{Table~\ref{tab:Table10}}, the S\_Then\_C approach outperforms the S\_Then\_C approach, particularly in F1 score, DSC, and HD, with statistically significant differences (p $<$ 0.05). These findings suggest that, for standard plane classification, leveraging features learned from the segmentation task via a frozen encoder and fine-tuning only the decoder yields better results. This supports our hypothesis that segmentation-guided representation learning facilitates more discriminative features for the classification task. Conversely, in segmentation tasks, introducing non-standard plane features into the encoder (as occurs when training classification first) can compromise the extraction of essential anatomical features, leading to a slight degradation in segmentation accuracy. 

In addition to the two-stage training approach "S\_Then\_C" adopted by most teams, the one-stage training approach employed by T1 and T3 has also achieved excellent results.  T1 integrated the three tasks of classification, segmentation, and measurement into a unified framework for training, achieving the overall first-place result. T3 combined classification and segmentation tasks within a single framework, securing the top rank in the biometry task. These results partially demonstrate the advantages of the one-stage training approach.

\subsection{Effects of Data Augmentation on Segmentation Performance}
In addition to the training approach, data augmentation methods are also critical aspects for enhancing network performance. Therefore, we investigated different data augmentation techniques on model segmentation performance. Elastic deformation and affine transformation (encompassing scaling, translation, shearing, and rotation) are employed to simulate the natural spatial variations and morphological changes that may occur in real-world imaging scenarios, enabling the model to adapt to diverse geometric distortions. Dropout, which randomly sets pixels in a rectangular region to 0, helps enhance the model's tolerance to partial information loss and prevents over-reliance on specific local features. Unsharp Mask and CLAHE (contrast-limited adaptive histogram equalization) focus on improving image quality: the former enhances contrast through unsharp masking to highlight subtle details, while the latter adjusts histogram distribution to mitigate over-enhancement of noise, ensuring more stable feature extraction. 

\textbf{Fig.~\ref{Figure9}} illustrates the effects of Elastic, Affine, Dropout, CLAHE, and Unsharp Mask augmentation methods on DSC (\textbf{Fig.~\ref{Figure9}A}), ASSD (\textbf{Fig.~\ref{Figure9}B}), and HD (\textbf{Fig.~\ref{Figure9}C}) metrics. Experimental results show that various augmentation methods improve segmentation performance to different degrees. Collectively, these methods comprehensively augment the dataset in terms of spatial transformations and intensity adjustments, fostering a more robust model capable of handling the variability inherent in practical applications.

\subsection{Effects of Model Architectures on Segmentation and Biometry Performance}
\begin{table*}[!ht]
\centering
\caption{Summary of the quantitative evaluation results of classification, segmentation and biometry tasks by 17 Deep Learning Methods (Including the Top 5 Methods in the Competition). \textbf{The best results for each metric are highlighted in bold.}}
\label{tab:Table11}
\fontsize{7}{8} \selectfont
\begin{tabular}{@{}r|cccc|ccc|cc@{}}
\toprule
\multirow{2}{*}{\textbf{Model}} & \multicolumn{4}{c|}{\textbf{Classification}} & \multicolumn{3}{c|}{\textbf{Segmentation}} & \multicolumn{2}{c}{\textbf{Biometry}} \\
\cmidrule(r){2-5} \cmidrule(lr){6-8} \cmidrule(l){9-10}
& \textbf{ACC}$\uparrow$ & \textbf{F1}$\uparrow$ & \textbf{AUC}$\uparrow$ & \textbf{MCC}$\uparrow$ & \textbf{DSC}$\uparrow$ & \textbf{ASSD}$\downarrow$ & \textbf{HD}$\downarrow$ & \textbf{$\Delta$AoP}$\downarrow$ & \textbf{$\Delta$HSD}$\downarrow$ \\
\midrule
\multirow{2}{*}{AccUnet (A15)} 
& 0.7078 & 0.7254 & 0.7172 & 0.3032 & 0.8217 & 16.7157 & 58.0059 & 13.2327 & 17.4524 \\
& (0.2261) & (0.2353) & (0.3243) & (0.3137) & (0.1285) & (13.4618) & (38.4260) & (13.2327) & (17.4524) \\
\hline
\multirow{2}{*}{Convformer (A12)} 
& 0.5701 & 0.3534 & 0.6996 & 0.2165 & 0.8863 & 11.6433 & 49.0574 & 16.8937 & 12.5160 \\
& (0.2696) & (0.3590) & (0.3375) & (0.3296) & (0.0712) & (7.9878) & (34.2426) & (16.8937) & (12.1931) \\
\hline
\multirow{2}{*}{CQUT-Smart (T4)} 
& 0.6319 & 0.6780 & 0.6225 & 0.2423 & 0.8535 & 11.0677 & 37.3815 & 10.7543 & 10.7867 \\
& (0.1900) & (0.2088) & (0.3028) & (0.2910) & (0.0596) & (5.1738) & (25.7057) & (10.0259) & (9.6300) \\
\hline
\multirow{2}{*}{Fatnet (A6)}  
& 0.5208 & 0.2427 & 0.6813 & 0.1412 & 0.9193 & 8.2690 & 35.0846 & 8.9003 & 11.0113 \\
& (0.2617) & (0.3169) & (0.3169) & (0.2494) & (0.0313) & (3.9062) & (24.9813) & (14.5212) & (15.7636) \\
\hline
\multirow{2}{*}{Ganjie (T1)}  
& \textbf{0.7441} & \textbf{0.7555} & \textbf{0.7802} & 0.3648 & 0.8475 & 12.9975 & 38.7210 & 10.4283 & 11.4523 \\
& (0.2224) & (0.2278) & (0.3355) & (0.3553) & (0.0638) & (20.2825) & (31.6393) & (11.0271) & (9.4196) \\
\hline
\multirow{2}{*}{H2former (A8)}  
& 0.7125 & 0.7462 & 0.7724 & 0.3186 & 0.9133 & 8.6165 & 33.7433 & 9.7218 & 10.1166 \\
& (0.2266) & (0.2289) & (0.3283) & (0.3246) & (0.0314) & (3.2991) & (17.7482) & (9.3986) & (9.0191) \\
\hline
\multirow{2}{*}{MambaUnet (A1)}  
& 0.7035 & 0.6880 & 0.7104 & 0.3338 & 0.9183 & 8.0547 & 29.0070 & 8.4009 & 8.2046 \\
& (0.2322) & (0.2726) & (0.3365) & (0.3480) & (0.0331) & (3.4261) & (14.6737) & (8.5773) & (8.2622) \\
\hline
\multirow{2}{*}{Missformer (A16)}  
& 0.6796 & 0.7451 & 0.7572 & 0.2418 & 0.8724 & 12.4940 & 45.2044 & 11.3457 & 13.0385 \\
& (0.2369) & (0.2106) & (0.3376) & (0.2976) & (0.0447) & (5.1525) & (19.3351) & (9.7412) & (10.9740) \\
\hline
\multirow{2}{*}{nkdinsdale95 (T5)}   
& 0.6789 & 0.7488 & 0.7571 & 0.2519 & 0.8169 & 14.3876 & 40.4391 & 15.3364 & 15.2268  \\
& (0.2386) & (0.2056) & (0.3355) & (0.3034) & (0.0906) & (15.7575) & (23.8167) & (16.4796) & (16.2635) \\
\hline
\multirow{2}{*}{nnUnet V2 (A7)} 
& - & - & - & - & 0.9021 & 9.8477 & 33.2985 & 8.5586 & 9.6310 \\
& (-) & (-) & (-) & (-) & (0.0354) & (3.9485) & (16.4069) & (8.4164) & (8.3313) \\
\hline
\multirow{2}{*}{Poolformer (A9)} 
& 0.6091 & 0.4350 & 0.6833 & 0.2399 & 0.9137 & 8.6425 & 32.7659 & 7.7549 & 10.1381 \\
& (0.2243) & (0.3259) & (0.3250) & (0.2968) & (0.0345) & (3.9000) & (17.4468) & (7.3980) & (9.4308) \\
\hline
\multirow{2}{*}{BioMedIA (T3)}  
& 0.6619 & 0.5422 & 0.7126 & 0.2145 & 0.8633 & 12.2203 & 41.1129 & 9.1557 & 11.6758  \\
& (0.2499) & (0.3740) & (0.3365) & (0.3200) & (0.0663) & (5.8450) & (19.8497) & (8.9648) & (18.6454) \\
\hline
\multirow{2}{*}{SAM (A2)} 
& - & - & - & - & 0.9187 & 8.2993 & \textbf{26.7800} & \textbf{6.4452} & \textbf{7.0359} \\
& (-) & (-) & (-) & (-) & (0.0289) & (3.4450) & (12.0206) & (5.5004) & (5.5097) \\
\hline
\multirow{2}{*}{SwinUnet (A3)} 
& 0.7238 & 0.6432 & 0.7724 & \textbf{0.4385} & 0.9162 & 8.2685 & 30.0146 & 8.1579 & 7.9375 \\
& (0.2559) & (0.3475) & (0.3409) & (0.3721) & (0.0319) & (3.5700) & (16.4006) & (6.2513) & (6.7444) \\
\hline
\multirow{2}{*}{TransUnet (A4)}  
& 0.5795 & 0.3695 & 0.7065 & 0.1648 & \textbf{0.9222} & \textbf{7.7874} & 28.1913 & 7.4143 & 8.6514 \\
& (0.2354) & (0.3579) & (0.3316) & (0.2687) & (0.0282) & (2.9104) & (14.8740) & (6.3333) & (6.9023) \\
\hline
\multirow{2}{*}{Upernet (A10)} 
& 0.6291 & 0.5589 & 0.6667 & 0.2972 & 0.9018 & 9.6442 & 40.2813 & 8.9304 & 10.6948 \\
& (0.2664) & (0.3448) & (0.3459) & (0.3673) & (0.0554) & (4.9391) & (23.5631) & (8.6169) & (10.2296) \\
\hline
\multirow{2}{*}{ViCBiC (T2)}
& 0.5670 & 0.6306 & 0.6868 & 0.1286 & 0.8857 & 9.4349 & 28.4152 & 9.4899 & 10.3878 \\
& (0.2453) & (0.2717) & (0.3363) & (0.2319) & (0.0476) & (8.5084) & (22.1607) & (9.7362) & (8.9781) \\ 
\bottomrule
\end{tabular}
\end{table*}

\begin{figure}[htbp]
	\centering
\includegraphics[width=0.48\textwidth]{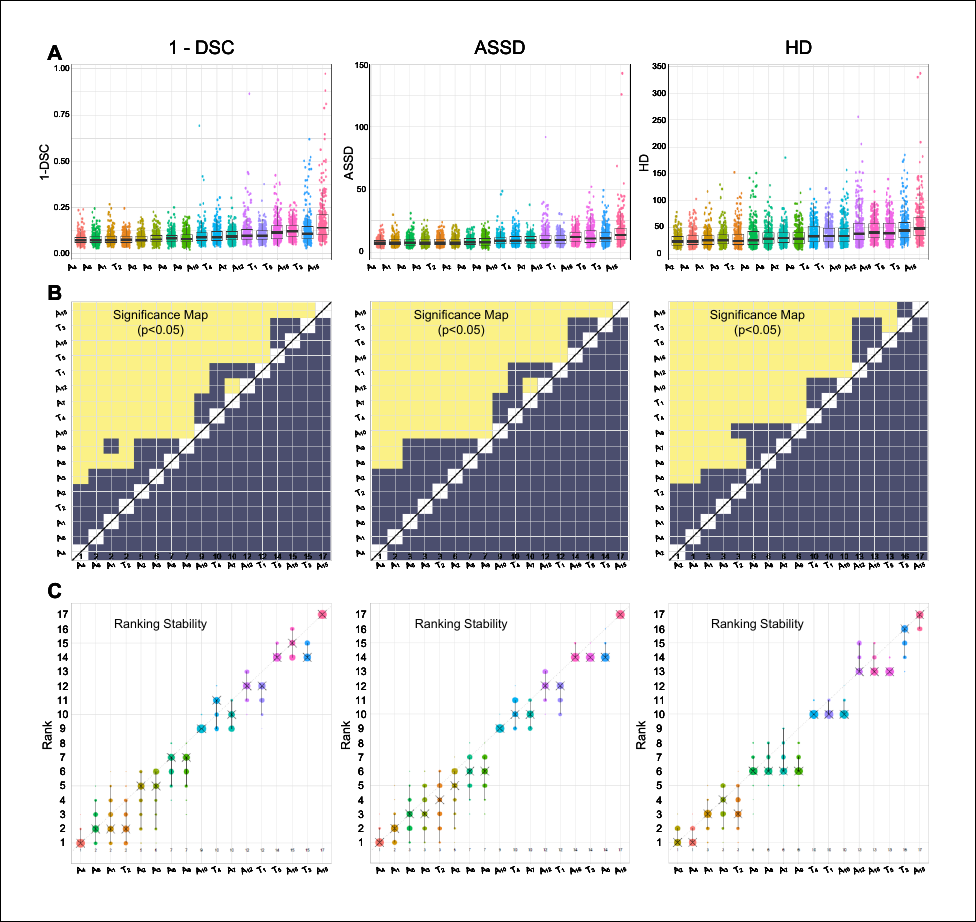}
\caption{Segmentation Performance of 17 Methods (including the Top 5 Methods in the Competition) with DSC (first column), ASSD (second column), and HD (third column). (A) Dot plots and boxplots visualize the distribution of evaluation metric data for each algorithm separately. (B) Significance maps visualize the results of significance testing. (C) Blob plots show the ranking stability. See Section 3.3.2 for details.}
\label{Figure10}
\end{figure}
\begin{figure*}[htbp]
	\centering
\includegraphics[width=0.95\textwidth]{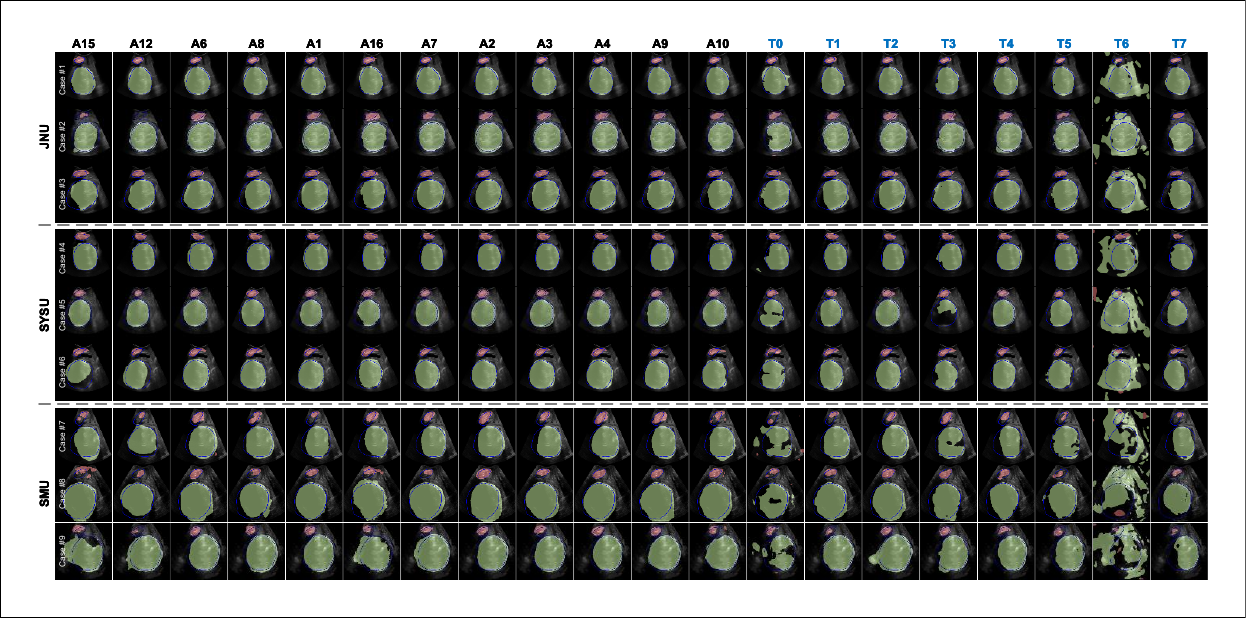}
\caption{Visualization of segmentation results for 20 methods on individual cases from JNU, SYSU, and SUM datasets. The red region denotes PS (predicted segmentation), the green region denotes FH (feature/contrast result), and the blue line represents the ground truth. The 20 methods involved are as follows:  Baseline (T0), Ganjie (T1), ViCBiC (T2), BioMedIA (T3), CQUT-Smart (T4), nkdinsdale95 (T5), HHL\_hotpot (T6), Serikbay (T7), Fatnet (A6), MambaUnet (A1), SwinUnet (A3), TransUnet (A4), Poolformer (A9), Upernet (A10), H2former (A8), AccUnet (A15), Convformer (A12), Missformer (A16), SAM (A2), and nnUnet V2 (A7).}
	\label{Figure11}
\end{figure*}
Beyond the influence of data augmentation on segmenation performance, the model architecture represents a critical aspect. To determine the optimal network architecture, we considered not only the top 5 methods (i.e., T1, T2, T3, T4, and T5) but also other models, including Fatnet (A6) \citep{2022FAT}, MambaUnet (A1) \citep{2024Mamba}, SwinUnet (A3) \citep{2021Swin}, TransUnet (A4) \citep{2024TransUNet}, Poolformer (A9) \citep{2025Edge}, Upernet (A10) \citep{2024ConvNeXt}, H2former (A8) \citep{he2023h2former}, AccUnet (A15) \citep{2023ACC}, Convformer (A12) \citep{lin2023convformer}, Missformer (A16) \citep{huang2022missformer},nnUnet V2 (A7) \citep{2024nnU} and SAM (A2) \citep{kirillov2023segment}. Note: SAM is integrated as an automatic segmentation backbone using bounding-box–based prompts. For the first frame of each video sequence, coarse foreground regions corresponding to the FH and PS are first obtained using the AoP-SAM \citep{zhou2025segment}. Connected component analysis is then applied to identify candidate anatomical regions, and the minimum enclosing bounding box of each component is used as the spatial prompt to initialize SAM segmentation. This design enables fully automatic prompt generation without manual interaction while providing anatomically interpretable region-of-interest localization. For subsequent frames, temporal prompt propagation is employed to ensure inter-frame consistency. Specifically, bounding-box prompts are propagated from the previous frame to the current frame using dense optical-flow–based motion estimation, followed by spatial smoothing and morphological dilation to accommodate fetal motion, maternal movement, and probe-induced displacement. Finally, segmentation outputs are post-processed using hole filling and small-component removal to enforce anatomical continuity and suppress spurious predictions, and the refined masks are subsequently passed to the biometry module for AoP and HSD computation.

For the segmentation task, TransUnet (A4) achieved the highest DSC, demonstrating comparable performance to Fatnet (A6), MambaUnet (A1), T2, and SAM (A2) (without statistical significance), but outperforming all other models. A4 also achieved the smallest ASSD, with performance comparable to A1, A6, A3, A2, and T2 (no significant differences), surpassing other architectures. SAM (A2) obtained the smallest HD, showing comparable results to A4, A1, A3, and T2, while outperforming the rest. Ranking stability analysis further reveals that A4 ranked first in both DSC and ASSD, while A2 and A4 tied for the top rank in HD. Notably, apart from T2, models A1, A2, and A4 also excel in segmentation performance (\textbf{Fig.~\ref{Figure10}} and \textbf{Table ~\ref{tab:Table11}}).

From the qualitative visualization results (\textbf{Fig.~\ref{Figure11}}), distinct failure patterns can be observed across different acquisition domains. For representative cases from the JNU dataset (i.e., Case \#1, \#2, and \#3), most submitted challenge methods achieved accurate delineation of both PS and FH regions, while T6 exhibited noticeable boundary leakage and partial mis-segmentation, indicating limited robustness to local intensity inhomogeneity and weak boundary constraints. These results suggest that, under relatively stable imaging conditions, most methods can reliably capture the major anatomical structures. In contrast, SYSU cases (i.e., Case \#4, \#5, and \#6) reveal more pronounced performance variability. Several participant methods (T0, T5, and T6) as well as generic backbone models (A12 and A15) demonstrated degraded segmentation quality, characterized by fragmented FH contours and inaccurate PS localization. This degradation is likely related to domain-specific variations in probe configurations and image contrast, which challenge models that primarily rely on global appearance cues without sufficient domain adaptation or shape regularization mechanisms. Meanwhile, other methods maintained relatively consistent predictions across SYSU cases, indicating improved robustness achieved through task-specific training strategies and anatomical priors. The SMU dataset (i.e., Case \#7, \#8, and \#9) presents the most challenging scenario. A larger subset of models failed to achieve satisfactory segmentation performance, particularly for the PS region. Specifically, methods including T0, T1, T3, T5, T6, T7 and several additional baselines (A4, A9, A10, A12, A15, and A16) suffered from severe under-segmentation or boundary displacement. These failures can be attributed to lower image quality, increased speckle noise, and reduced tissue contrast, which exacerbate ambiguity at anatomical interfaces and expose the limitations of conventional convolutional or transformer-based architectures without explicit boundary modeling or uncertainty-aware mechanisms. Among all evaluated approaches, the additional model A1 consistently produced accurate and stable segmentation results across all three datasets, demonstrating superior cross-domain generalization capability. This advantage is likely related to its enhanced long-range dependency modeling and sequence-aware representation learning, which better capture global anatomical context while preserving fine-grained structural details. Overall, these qualitative observations are consistent with the quantitative rankings reported in \textbf{Table~\ref{tab:Table11}} and further highlight the importance of domain robustness and boundary-aware modeling strategies for reliable intrapartum ultrasound segmentation.

\begin{figure}[htbp]
	\centering
\includegraphics[width=0.48\textwidth]{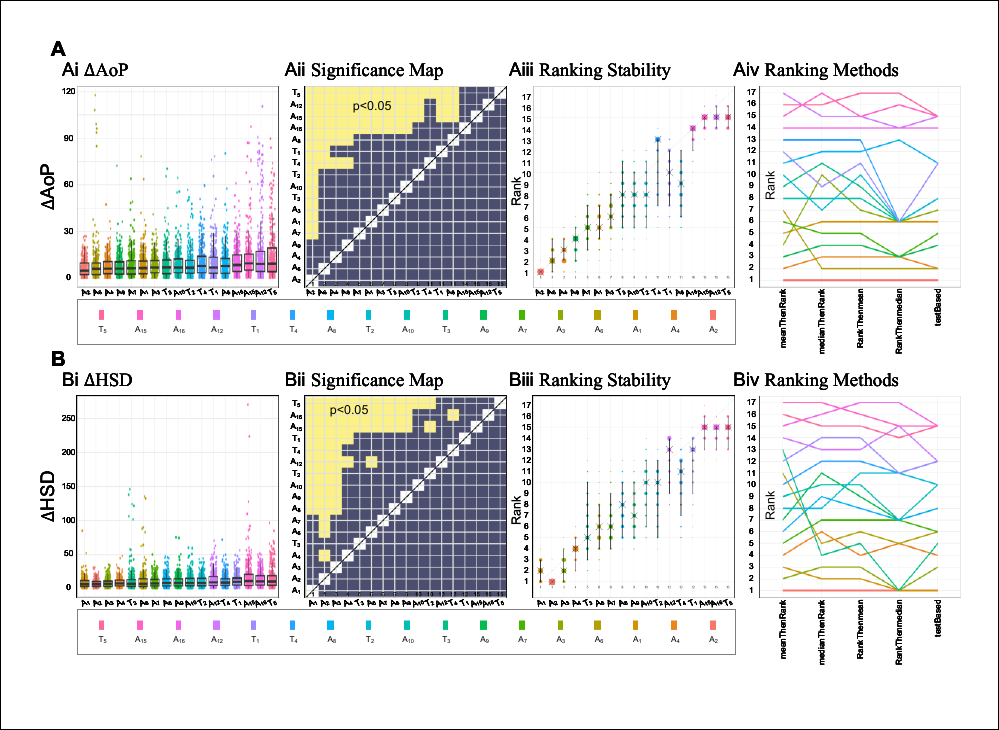}
\caption{Biometry Performance of 17 Deep Learning Methods (Including the Top 5 Methods in the Competition) for $\Delta$AoP (A) and $\Delta$HSD (B).  (Ai/Bi) Dot plots and boxplots: Visualize the distribution of evaluation metric data for each algorithm separately. (Aii/Bii) Significance maps visualize the results of significance testing. (Aiii/Biii) Blob plots show the ranking stability.  (Aiv/Biv) Line plots demonstrate the ranking robustness of each algorithm across different ranking methods.}
	\label{Figure12}
\end{figure}

For the biometry task, we further analyzed the performance of 17 methods in ultrasonic parameter measurement. Experimental results show that for AoP measurement, A2 achieved the smallest $\Delta$AoP, demonstrating comparable performance to A4, A6, and A9 (without statistical significance) but outperforming other methods. A2 ranked first and consistently stayed at the top across five ranking schemes. For HSD measurement, A1 achieved the smallest $\Delta$HSD, with performance comparable to A2, A3, A4, T3, A6, and A7 (no significant differences) but superior to others. While A1's top ranking was inconsistent, A2 consistently ranked first across all five ranking schemes (\textbf{Fig.~\ref{Figure12}}). 

Thus, these results indicate that from a model architecture perspective, A2 serves as the benchmark model for the IUGC challenge, as it achieved optimal results in both segmentation and measurement tasks.

\subsection{Effects of Post-processing Methods on Biometry Performance}
The ultimate goal of this challenge is to achieve automated measurement of ultrasonic parameters, where existing methods rely on post-processing of segmentation outputs. Most participating teams adopted the same post-processing strategy as Team T0, which directly determines coordinate points for ultrasonic parameter calculation without elliptical fitting of segmentation results. In contrast, several teams derived coordinate points by performing elliptical fitting on the segmented regions. To systematically investigate the impact of these post-processing strategies on measurement accuracy, four processing schemes were designed: no elliptical fitting for PS/FH (w/o), elliptical fitting for both PS/FH (w/e\_PS\_FH), elliptical fitting for FH only (w/e\_FH), and elliptical fitting for PS only (w/e\_PS). We quantitatively compared the impact of these four schemes on ultrasound parameter estimation performance.

\begin{figure}[htbp]
	\centering
\includegraphics[width=0.4\textwidth]{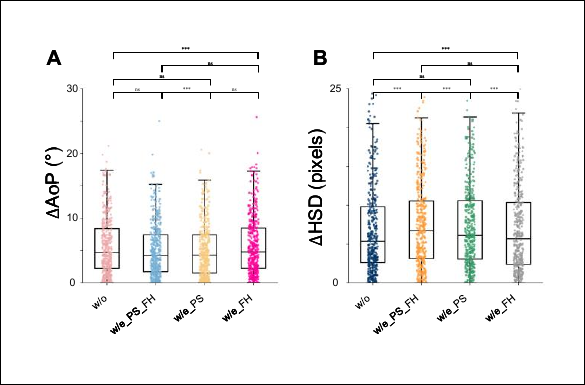}
\caption{Effects of Different Post-Processing Methods on Biometric Parameter Measurement Results Based on $\Delta$AoP (A) and $\Delta$HSD (B). w/o: no elliptical fitting applied to both segmentation results of PS and FH; w/e\_PS\_FH: elliptical fitting applied to both segmentation results of PS and FH; w/e\_PS: elliptical fitting applied only to PS segmentation results; and w/e\_FH: elliptical fitting applied only to FH segmentation results. See Section 3.3.2 for details.}
	\label{Figure13}
\end{figure}

Experimental results show that for AoP measurement, w/e\_PS\_FH achieves the best average performance, although the improvement over w/o is marginal. In contrast, for HSD measurement, w/o consistently outperforms w/e\_PS\_FH with statistically significant differences (\textbf{Fig.~\ref{Figure13}}). Notably, the w/o strategy yields the most stable overall performance across both AoP and HSD measurements.

This phenomenon can be explained by the boundary-driven nature of our biometry pipeline. The key landmarks used to compute AoP and HSD are extracted directly from the predicted PS and FH contours. Consequently, segmentation inaccuracies propagate to both landmark localization and downstream measurements. In particular, spurious segmentations (false-positive regions), boundary leakage, and contour discontinuities may introduce outlier points or erroneous curvature extrema along the contour, which can shift the detected tangent or contact points and bias AoP and HSD estimation. When elliptical fitting is applied, these effects may be further amplified because the fitting step enforces a global geometric model on potentially noisy and incomplete contours. Local outliers caused by spurious regions can substantially alter the fitted ellipse parameters (including center location, major and minor axes, and rotation angle), leading to a global displacement of the fitted contour. As a result, landmarks computed on the fitted ellipse may deviate from the true anatomical interface, introducing systematic measurement bias rather than stabilizing the estimation.

This behavior is particularly pronounced in challenging intrapartum ultrasound frames characterized by low contrast, motion-induced blur, and partial anatomical occlusions, where segmentation uncertainty is higher. Under such conditions, directly using raw contour geometry (w/o) avoids imposing overly restrictive shape assumptions and better preserves local anatomical fidelity, thereby yielding more robust measurement performance. These findings suggest that classical geometric regularization techniques, such as ellipse fitting, may not be universally beneficial in dynamic and noisy clinical imaging scenarios and should be applied with caution or combined with segmentation quality assessment and uncertainty-aware filtering strategies.

\subsection{Comparison with Previous Studies}
Although several prior studies have reported low AoP measurement errors on single-center datasets (e.g., $\Delta$AoP = $3.90^\circ$ in \citep{lu2022multitask},  $\Delta$AoP = $2.66^\circ$ in \citep{Chen2024Ultrasound} and $\Delta$AoP = $3.81^\circ$ in \citep{bai2026iugc} , as summarized in Table~\ref{tab:Table2}), the best-performing method in the IUGC multi-center benchmark achieved $\Delta$AoP = $9.16^\circ$. This discrepancy reflects the substantially increased difficulty and clinical realism of multi-center evaluation.

First, many prior works evaluate AoP measurement under constrained acquisition conditions, typically using manually or automatically pre-selected standard planes where anatomical landmarks are well aligned and image quality is optimized \citep{lu2022multitask,zhou2020automatic,2022OC05,bai2022framework}. Such settings reduce geometric variability and localization ambiguity. In contrast, the IUGC benchmark evaluates end-to-end performance on unconstrained intrapartum ultrasound video streams, which include non-standard views, probe motion, fetal movement, partial occlusions, and variable acquisition angles. This setting more closely reflects real clinical workflows and introduces significantly higher spatial and temporal uncertainty.

Second, several previous studies adopt semi-automatic or user-assisted measurement pipelines, where expert input is involved in region selection     \citep{conversano2025automated,F2017Automatic,Angeli2020New}. These human-in-the-loop designs inherently benefit from operator guidance and reduce measurement ambiguity. By contrast, the IUGC challenge enforces fully automatic inference without manual intervention, eliminating operator bias and imposing stricter deployment-level constraints.

Third, most single-center benchmarks are based on homogeneous imaging environments with consistent devices, acquisition protocols, and patient populations. Models trained and evaluated under such conditions may implicitly overfit institution-specific appearance patterns and noise characteristics. In contrast, the IUGC dataset integrates multiple hospitals, ultrasound systems, operators, and patient cohorts, introducing realistic domain shifts that directly affect segmentation robustness, landmark localization stability, and downstream biometric accuracy.

\subsection{Clinical Impact}
\begin{figure}[htbp]
	\centering
\includegraphics[width=0.45\textwidth]{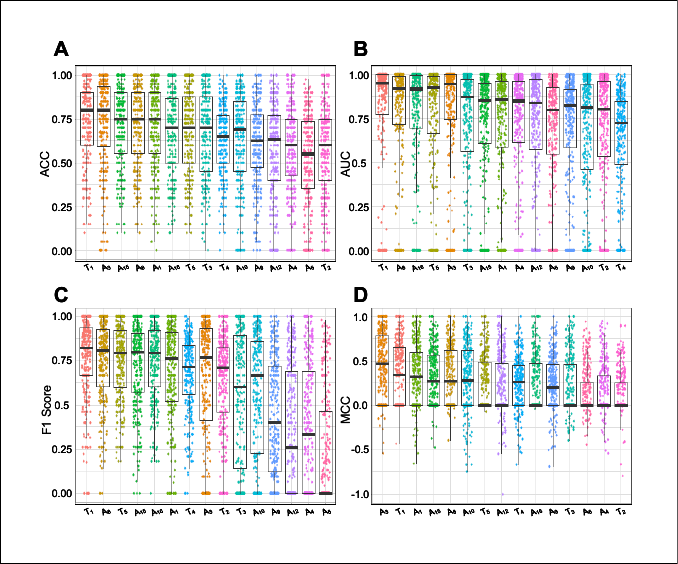}
\caption{Correlation between ultrasound parameters across different source datasets (i.e., JNU, SYSU, SMU and All representing the combined data from all three sources).}
	\label{Figure14}
\end{figure}

Both AoP and HSD are distinct indicators for quantifying fetal head station, and they exhibit a strong linear relationship (\textbf{Fig.~\ref{Figure14}}). On the one hand, the consistency of measurement indices can be evaluated based on the correlation between the two; on the other hand, the combined use of these two indices for assessing fetal head station can enhance the reliability of automated measurement results. The results of the IUGC Challenge and further in-depth analysis show that video-based end-to-end multi-tasking approaches (e.g., T1) can achieve optimal outcomes. Additionally, the application of ultrasound-specific data augmentation techniques, the integration of pre-trained networks (i.e., SAM, ImageNet), and post-processing methods using non-elliptical fitting can improve algorithm accuracy. To render these technologies clinically applicable, large-scale clinical trials are still required to validate and optimize them.
\subsection{Limitation and Future Prospects}
\begin{figure}[htbp]
	\centering
\includegraphics[width=0.3\textwidth]{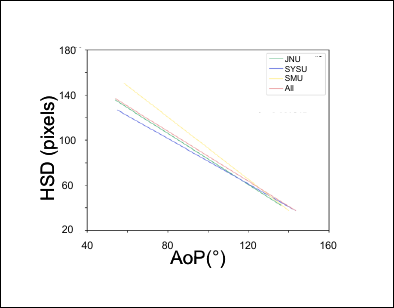}
\caption{Classification Performance of 17 Deep Learning Methods (including the Top 5 Methods in the Competition). A-D are Dot plots and boxplots of ACC, AUC, F1 score and MCC, respectively.}
	\label{Figure15}
\end{figure}

This study encompasses several notable limitations. Firstly, in the IUGC's three cascaded tasks, the accuracy of the initial classification stage serves as a critical foundation that directly affects the performance of subsequent segmentation and biometry modules. However, a significant shortcoming is observed: none of the submitted methods achieved an accuracy exceeding 80\% for standard-plane identification (\textbf{Fig.~\ref{Figure15}} and \textbf{Table~\ref{tab:Table11}}), which remains substantially lower than the performance of expert manual identification (above 86\%, \textbf{Table~\ref{tab:Table4}}). This gap highlights the intrinsic difficulty of automatic standard-plane recognition and the urgent need for further methodological advances. 

To better understand this limitation, we analyzed the confusion matrices of all evaluated methods (\textbf{Supplementary Fig. S2}). The test set consists of 8,665 frames extracted from 300 ultrasound videos, with a moderately imbalanced class distribution of 4,822 standard-plane and 3,843 non-standard-plane images. Across both challenge submissions (T0--T7) and additional baseline methods, a consistent error pattern is observed: most models exhibit substantially higher false-positive rates than false-negative rates, indicating a tendency to misclassify visually similar non-standard frames as standard planes. This behavior is primarily attributed to the low inter-class separability between standard and non-standard views. Many non-standard frames contain partial anatomical structures that closely resemble target planes, leading to ambiguous visual cues near the decision boundary. Meanwhile, standard-plane images themselves exhibit large intra-class variability caused by probe angle variation, speckle noise, acoustic shadowing, and fetal motion, further increasing classification difficulty. In addition, although the dataset is video-based, where standard and non-standard planes typically appear in temporally continuous segments, most evaluated methods rely on independent frame-level classification without explicitly exploiting temporal consistency or motion information. This limitation increases sensitivity to transient artifacts and contributes to misclassification around plane transition regions.

Secondly, despite the challenge providing both labeled data and a substantial amount of unlabeled data, only Team T3 among the eight participating teams leveraged unlabeled samples to improve model performance. This underutilization highlights the unrealized potential of semi-supervised and self-supervised learning strategies, which remain promising directions for future benchmark development. From a methodological perspective, most participating teams adopted image-based two-stage pipelines. In contrast, Team T1 employed a video-based end-to-end framework, which achieved the best overall performance in the challenge, suggesting that further exploration of temporally-aware and sequence-level learning paradigms may yield additional performance gains.

Thirdly, while the T1 and A1 (SAM-based) methods demonstrated strong performance, their large parameter counts introduce practical limitations for real-time deployment. These high-capacity models are not well suited for edge-device inference in time-critical clinical settings, motivating future research on lightweight architectures, model compression, and hardware-efficient deployment strategies.

Fourthly, although the benchmark integrates multi-center data acquired from three ultrasound devices, the overall source composition remains limited and imbalanced. Performance degradation on minority-source subsets is more pronounced for classification and downstream biometry tasks, whereas segmentation performance remains comparatively stable. This imbalance may introduce cohort-level bias by over-representing dominant acquisition domains while under-representing challenging clinical scenarios, potentially leading to overly optimistic aggregated performance estimates. To address this issue, future benchmark iterations will prioritize expanding device diversity and geographic coverage, constructing balanced evaluation subsets, and incorporating source-aware reporting strategies, such as macro-averaged metrics and per-source performance breakdowns, to more reliably assess domain generalization capability.

Finally, although the challenge design closely reflects real clinical workflows by integrating classification, segmentation, and biometry into a unified cascaded pipeline, this multi-stage framework is inherently vulnerable to error propagation and increased computational overhead. Exploring alternative end-to-end keypoint detection or direct regression-based measurement frameworks may offer a promising direction to reduce cumulative errors while improving computational efficiency and deployment feasibility.
\section{Conclusion}\label{sec:con}
The IUGC Challenge represents a pivotal advancement in leveraging deep learning for automated intrapartum ultrasound analysis, introducing a multi-task framework and the largest multi-center video dataset (774 videos) to date for concurrent standard plane classification, fetal head-pubic symphysis segmentation, and biometric measurement (HSD and AoP). Participating teams demonstrated diverse strategies, from Video Swin Transformer-based end-to-end video processing to hybrid UNet-Transformer architectures, achieving notable performance: T1 excelled in classification (ACC=0.7441), T2 in segmentation (DSC=0.8857), and T3 in biometry ($\Delta$AoP=9.1557, $\Delta$HSD=10.1381). The challenge highlighted the efficacy of multi-source data, pre-trained models (e.g., SAM), and task-specific augmentations, while publicly releasing the dataset and solutions to foster field progress. Notably, all methods exploited complementary information across tasks, but key limitations persist: standard plane identification accuracy lagged behind manual annotation ($<$80\% vs. $>$86\%), unlabeled data remained underutilized, and multi-task error accumulation plagued end-to-end pipelines. Model complexity also hindered real-time edge deployment, with SAM showing top performance but high parameter counts. Looking ahead, future research should prioritize improving standard plane detection robustness, integrating semi-supervised learning for unlabeled data, developing lightweight end-to-end video frameworks, and mitigating error propagation via joint registration-measurement pipelines. These advancements will bridge the gap toward clinical translation, enhancing labor assessment efficiency and reducing reliance on manual analysis globally.

\section*{CRediT authorship contribution statement} Jieyun Bai contributed to Conceptualization, Formal Analysis, Investigation, Methodology, Software, Validation, Visualization, Writing–Original Draft, and Writing–Review \& Editing. Zihao Zhou was involved in Formal Analysis, Investigation, Methodology, Software, Validation, Visualization, and Writing–Review \& Editing. Yitong Tang participated in Formal Analysis, Investigation, Methodology, Software, and Writing–Review \& Editing. Saad Slimani, Victor M. Campello, Isaac Khobo, and Bernard Ohene-Botwe contributed to Conceptualization and Writing–Review \& Editing. Yuxin Huang, Zhenyan Han, Hongying Hou, and Zheng Zheng handled Conceptualization, Validation, and Writing–Review \& Editing. Gongning Luo, Dong Ni, Shuo Li, and Yaosheng Lu were responsible for Conceptualization, Funding Acquisition, and Writing–Review \& Editing. Karim Lekadir managed Funding Acquisition, Project Administration, Resources, Supervision, Writing–Review \& Editing, and Validation. All remaining authors contributed to Methodology, Software, and Writing–Review \& Editing. Specific groups included: Ganjie (Jie Gan, Zhuonan Liang, Lisa B. Mcguire, Jillian L. Clarke, Weidong Cai, Jacqueline Spurway), ViCBiC (Yubo Tan, Shiye Wang, Wenda Shen, Wangwang Yu, Yihao Li, Philippe Zhang, Weili Jiang, Yongjie Li), BioMedIA (Salem Muhsin Ali Binqahal Al Nasi, Arsen Abzhanov, Numan Saeed, Mohammad Yaqub), CQUT-Smart (Zunhui Xia, Hongxing Li, Libin Lan), nkdinsdale95 (Jayroop Ramesh, Valentin Bacher, Mark Eid, Hoda Kalabizadeh, Christian Rupprecht, Ana I. L. Namburete, Pak-Hei Yeung, Madeleine K. Wyburd, Nicola K. Dinsdale), HHL\_hotpot (Zicheng Hu, Nana Liu, Yian Deng, Wei Hu, Cong Tan, Wenfeng Zhang, Mai Tuyet Nhi), Serikbay (Assanali Serikbey, Jiankai Li, Sung-Liang Chen), Baseline (Zihao Zhou, Yaosheng Lu, Yitong Tang, Jieyun Bai), SAM (Marawan Elbatel, Robert Martí, Xiaomeng Li), and nnUnet v2 (Gregor Koehler, Raphael Stock, Klaus Maier-Hein).
\section*{Funding}
This work was supported by Guangzhou Science and Technology Planning Project (2025B03J0127), the National Institute of Hospital Administration (YLXX24AIA006), Sichuan Provincial Cross-Regional Innovation Cooperation Project (2025YFHZ0326), and Key Research and Development Program of Guangxi Province (2023AB22074 and 2024AB04027), 
Natural Science Foundation of Guangdong Province under Grant (2024A1515011886 and 2023A1515012833), High Expert Program (H20240205), Guangzhou Municipal Science and Technology Bureau Guangzhou Key Research and Development Program under Grant 2024B03J1283 and 2024B03J1289, Guangzhou Science and Technology Planning Project under Grant 2023B03J1297, China Scholarship Council under Grant 202206785002, National Key Research and Development Project (2019YFC0120100, 2019YFC0121907, and 2019YFC0121904), National Natural Science Foundation of China under Grant 61901192, and the European Research Council (ERC). The ERC, under the European Union’s Horizon Europe research and innovation programme (AIMIX project - Grant Agreement No. 101044779).(Corresponding author: Jieyun Bai and Yuxin Lasse Hansen.)
\section*{Declaration of Generative AI and AI-assisted technologies in the writing process} 
ChatGPT was utilized to enhance language clarity and readability. The authors reviewed, edited, and assume full responsibility for all content.  
\section*{Declaration of competing interest} The authors have declared no conflict of interest. 
\section*{Acknowledgements}
We thank Guangzhou Lianyin Medical Technology Co., Ltd., which sponsored us, and all individuals involved in data annotation.
\section*{Data availability} Links to the data and code repositories: 

\href{https://github.com/maskoffs/IUGC2024} {\textbf{All challenge-developed algorithms}} (GitHub).

\href{https://www.youtube.com/watch?v=xntGzr70KX8} {\textbf{The Recording of the IUGC session}} (YouTube).
\appendix 
\section{Supplementary Information}
\textbf{Participation Rules.} To guarantee impartiality and transparency in the IUGC Challenge, organizers, data providers, and contributors were barred from taking part in the competition. This restriction stemmed from the fact that data providers and organizers had access to the dataset, including the ground truth of the test set. Nevertheless, individuals affiliated with the organizing institutions were permitted to participate; however, they were ineligible for prizes and would not be included in the leaderboard. Teams that won prizes were required to provide their source codes, present their methodologies at MICCAI 2024, sign all required prize acceptance documents, and submit a detailed paper delineating their methodologies. Additionally, participants undertook to cite both the data challenge paper and this challenge overview paper in their subsequent publications, whether scientific or non-scientific in nature. The outcomes and rankings of the challenge were publicly released upon the completion of the testing phase. Note: Guangzhou Lian-Med Technologies Co., Ltd. is the principal sponsor of the IUGC. Only the organizers and technical groups have access to test case labels.

\textbf{Prize Policies.} The teams that rank first in each task and the top 7 in the comprehensive ranking will receive rewards. The team ranking first in the classification task is T1 (awarded a bonus of 1,000 RMB), the team ranking first in the segmentation task is T2 (awarded a bonus of 1,000 RMB), and the team ranking first in the biological parameter measurement task is T3 (awarded a bonus of 1,000 RMB). The top seven teams in the comprehensive ranking are T1 (awarded a bonus of 3,000 RMB and a certificate), T2 (awarded a bonus of 2,000 RMB and a certificate), T3 (awarded a bonus of 1,000 RMB and a certificate), T4 (certificate), T5 (certificate), T6 (certificate), and T7 (certificate).

\bibliographystyle{model2-names.bst}\biboptions{authoryear}
\bibliography{refs}
\end{document}